\documentclass{article}

\usepackage{arxiv}
\usepackage[utf8]{inputenc} 
\usepackage[T1]{fontenc}    
\usepackage{hyperref}       
\usepackage{url}            
\usepackage{booktabs}       
\usepackage{amsfonts}       
\usepackage{nicefrac}       
\usepackage{microtype}      
\usepackage{lipsum}		
\usepackage{graphicx}
\usepackage{natbib}
\usepackage{doi}
\newtheorem{prop}{Proposition}
\usepackage{amsmath, calc}
\usepackage{url}

\usepackage[ruled,vlined]{algorithm2e}
  
\makeatletter
\let\@fnsymbol\@arabic
\makeatother
\title{Fast and Interpretable Consensus Clustering via Minipatch Learning}

\author{ Luqin Gan
\thanks{Department of Statistics, Rice University, Houston, Texas 77005}

\And Genevera I. Allen
\thanks{Departments of Electrical and Computer Engineering, Statistics, and Computer Science, Rice University, Houston, Texas, 77005}
\textsuperscript{ ,}
\thanks{Duncan Neurological Research Institute, Baylor College of Medicine, Houston, Texas, 77030}
}
\begin{document}
\maketitle

\begin{abstract}
Consensus clustering has been widely used in bioinformatics and other applications to improve the accuracy, stability and reliability of clustering results.  This approach ensembles cluster co-occurrences from multiple clustering runs on subsampled observations.  For application to large-scale bioinformatics data, such as to discover cell types from single-cell sequencing data, for example, consensus clustering has two significant drawbacks: (i) computational inefficiency due to repeatedly applying clustering algorithms, and (ii) lack of interpretability into the important features for differentiating clusters. In this paper, we address these two challenges by developing IMPACC: Interpretable MiniPatch Adaptive Consensus Clustering.  Our approach adopts three major innovations.  We ensemble cluster co-occurrences from tiny subsets of both observations and features, termed minipatches, thus dramatically reducing computation time.  Additionally, we develop adaptive sampling schemes for observations, which result in both improved reliability and computational savings, as well as adaptive sampling schemes of features, which leads to interpretable solutions by quickly learning the most relevant features that differentiate clusters. We study our approach on synthetic data and a variety of real large-scale bioinformatics data sets; results show that our approach not only yields more accurate and interpretable cluster solutions, but it also substantially improves computational efficiency compared to standard consensus clustering approaches.
\end{abstract}

\section{Introduction}\label{sec:introduction}

Consensus clustering is a widely used unsupervised ensemble method in the domains of bioinformatics, pattern recognition, image processing, and network analysis, among others. This method often outperforms conventional clustering algorithms by ensembling cluster co-occurrences from multiple clustering runs on subsampled observations \citep{ghaemi2009survey}. However, consensus clustering has many drawbacks when dealing with large data sets typical in bioinformatics.  These include computational inefficiency due to repeated clustering of very large data on multiple subsamples, degraded clustering accuracy due to high sensitivity to irrelevant features, as well as a lack of interpretability.  Consider, for example, the task of discovering cell types from single-cell RNA sequencing data.  This data often contains tens-of-thousands of cells and genes, making consensus clustering computationally prohibitive.  Additionally, only a small number of genes are typically responsible for differentiating cell types; consensus clustering considers all features and provides no interpretation of which features or genes may be important.  Inspired by these challenges for large-scale bioinformatics data, we propose a novel approach to consensus clustering that utilizes tiny subsamples or minipatches as well as adaptive sampling schemes to speed computation and learn important features.

\subsection{Related Work}\label{sec:realtedwork}

Several types of consensus functions in ensemble clustering have been proposed, including co-association based function \citep{fred2001finding,fred2002data,kellam2001comparing,azimi2006clustering}, hyper-graph partitioning \citep{strehl2002cluster,ng2001spectral,karypis1998fast}, relabeling and voting approach \citep{dudoit2003bagging,fischer2003path,fischer2003bagging}, mixture model \citep{topchy2004mixture,topchy2005clustering,analoui2007solving}, and mutual information \citep{luo2006combining,topchy2003combining,azimi2007new}. Co-association based function, such as consensus clustering, is faster in convergence and is more applicable to large-scale bioinformatics data sets.  Our approach is based on consensus clustering, whose concept is straightforward. In order to achieve evidence accumulation, a consensus matrix is constructed from pairwise cluster co-occurrence, ranging in $[0,1]$. It is later regarded as a similarity matrix of the observations to obtain the final clustering results \citep{fred2005combining}. Closely related to our work, numerous variants of consensus clustering with adaptive subsampling strategies on observations have been proposed. For instance, \citet{duarte2012adaptive} update the sampling weights of objects with their degrees of confidence, which are subtracted by clustering the consensus matrix; \citet{parvin2013data} compute sampling weights by the uncertainty of object assignments based on consensus indices' distances to $0.5$; and \citet{topchy2004adaptive} adaptively subsample objects according to the consistency of clustering assignments in previous iterations. Besides adaptive sampling, \citet{ren2017weighted} overweight the observations with high confusion, and assigns the one-shot weights to obtain final clustering results. However, the existing sampling schemes focus on observations only and do not take feature relevance in to consideration. So these methods show inferior performance in application to sparse data sets, where only a small set of features can significantly influence cluster assignments. 

Many clustering methods and pipelines have been proposed that specifically focus on single cell RNA-seq data \citep{kiselev2017sc3,yang2019safe,satija2015spatial,wolf2018scanpy,trapnell2014dynamics}. A popular approach, SC3 \citep{kiselev2017sc3}, employs consensus clustering by applying dimension reduction to the subsampled data and then applying K-means. \citet{satija2015spatial} integrate dropout imputation and dimension reduction with a graph-based clustering algorithm. Another widely used and simple approach is to conduct tSNE dimension reduction followed by K-Means clustering \citep{kiselev2019challenges}. Many have discussed the computational challenge of clustering large-scale single-cell sequencing data \citep{kiselev2019challenges} and have sought to address this via dimension reduction.  But clustering based on dimension reduced data is no longer directly interpretable; that is, one cannot determine which genes are directly responsible for differentiating cell type clusters.  The motivation of our approach is not only to develop a fast computational approach, but also to develop a method which has built-in feature interpretability to discover deferentially expressed genes.

 A series of clustering algorithms have been proposed to add insights on feature importance. Some clustering algorithms conduct sparse feature selection through regularization within clustering algorithms. For example, sparse K-Means (sparseKM), sparse hierarchical clustering (spaeseHC) \citep{witten2010framework} and sparse convex clustering \citep{wang2018sparse,wang2021integrative} facilitate feature selection by solving a lasso type optimization problem. But this type of sparse clustering algorithm is often slow and highly sensitive to hyper-parameter choices; thus they face maybe computational challenges for large data. Another class of methods ranks features by their influence to results. The resulting sensitivity to the changes of one feature can be measured by the difference of silhouette widths of clustering results \citep{yu2019ensemble}, difference of the entropy of consensus matrices \citep{dash2000feature}, or consistency of graph spectrum \citep{zhao2007spectral}. However, feature ranking methods have to measure the importance of each feature separately, which lead to extremely high computational cost. Additionally, \citet{liu2018feature} propose a post-hoc feature selection method which solves an optimization problem to determine important features within the regular consensus clustering algorithm; however, this approach suffers from major computational hurdles for large data. Therefore, we are motivated to propose an extension of consensus clustering to dramatically improve clustering accuracy, provide model interpretability, and simultaneously ease the computational burden, by incorporating innovative adaptive sampling schemes on both features and observations with minipatch learning.

\subsection{Contributions}

In this paper, we propose a novel methodology as an extension of consensus clustering, which demonstrates major advantages in large scale bioinformatics data sets. Specifically, we seek to improve computational efficiency, provide interpretability in terms of feature importance, and at the same time improve clustering accuracy. We achieve these goals by leveraging the idea of minipatch learning \citep{yao2020feature, yao2021minipatch, toghani2021mp} which is an ensemble of learners trained on tiny subsamples of both observations and features.  Compared to only subsampling observations in existing consensus clustering ensembles, by learning on many tiny data sets, our approach offers dramatic computational savings.  In addition, we develop novel adaptive sampling schemes for both observations and features to concentrate learning on observations with uncertain cluster assignments and on features which are most important for separating clusters.  This provides inherent interpretations for consensus clustering and also further improves computational efficiency of the learning process.  We test our novel methods and compare them to existing approaches through extensive simulations and four large real-data case studies from bioinformatics and imaging.  Our results show major computational gains with our run time on the same order as that of hierarchical clustering as well as improved clustering accuracy, feature selection performance, and interpretability.

\section{Minipatch Consensus Clustering}\label{sec:methods}

Let $\boldsymbol{X}\in \mathbb{R}^{N\times M}$ be the data matrix of interest, with $M$ features measured over $N$ observations. $\boldsymbol{x_i}\in \mathbb{R}^{M}$ is the $M$-dimensional feature vector observed for sample $i$.  We assume that the observations can be separated into $K$ non-overlapping and exhaustive clusters; our goal is to find these clusters.  We propose to extend popular consensus clustering techniques \citep{monti2003consensus} to be able to more accurately and computationally efficiently detect clusters in high-dimensional noisy data common in bioinformatics \citep{hayes2006gene,verhaak2010integrated}.   We also seek ways to ensure our clusters are interpretable through feature selection.  To this end, we propose a number of innovations and improvements to consensus clustering outlined in our Minipatch Consensus Clustering framework in Algorithm~\ref{algo:rand}.  Similar to consensus clustering, our approach repeatedly subsamples the data, applies clustering, and records the $N \times N$ co-cluster membership matrix, $\boldsymbol{\mathcal{V}}$.  It then ensembles all the co-cluster membership information together into the $N \times N$ consensus matrix $\boldsymbol{\mathcal{S}}$.  This consensus matrix takes values in $[0,1]$ indicating the proportion of times two observations are clustered together; it can be regarded as a similarity matrix for the observations.  A perfect consensus matrix includes only entries of $0$ or $1$, where observations are always assigned to the same clusters; values in between indicate the (un)reliability of cluster assignments for each observation.  To obtain final cluster assignments, one can cluster the estimated consensus matrix, which typically yields more accurate clusters than applying standard, non-ensembled clustering algorithms \citep{ghaemi2009survey}.

While the core of our approach is identical to that of consensus clustering, we offer three major methodological innovations in Steps 1 and 2 of Algorithm~\ref{algo:rand} that yield dramatically faster, more accurate, and interpretable results.  Our first innovation is building cluster ensembles based on tiny subsets (typically 10\% or less) of both observations and features termed minipatches \citep{yao2020feature, yao2021minipatch, toghani2021mp}.  Note that existing consensus clustering approaches form ensembles by subsampling typically 80\% of observations and all the features for each ensemble member \citep{wilkerson2010consensusclusterplus}.  For large-scale bioinformatics data where the number of observations and features could be in the tens-of-thousands, repeated clustering of this large data is a major computational burden.  Instead, our approach termed Minipatch Consensus Clustering (MPCC) subsamples a tiny fraction of both observations and features and hence has obvious computational advantages:
\begin{prop}\label{le:tc}  
The computational complexity of MPCC in Algorithm~\ref{algo:rand} is $O(mn^2T+N^2)$, where $T$ is the total number of minipatches.
\end{prop}
Since $m$ and $n$ are very small, the dominating term is the $N^2$ computations required to update the consensus matrix.  This compares very favorably to existing consensus clustering approaches which if the default of 80\% of observations are subsampled in each run, then the time complexity is $O(MN^2T)$, which can be very slow for both large $N$ and large $M$ datasets.  On the other hand, our method is comparable in complexity to hierarchical clustering which is also $O(N^2)$ \citep{murtagh1983survey}, but is perhaps slower than K-Means which is $O(N)$ \citep{pakhira2014linear}. The proof of Proposition~\ref{le:tc} is given in Appendix~A.

While MPCC offers dramatic computational improvements over standard consensus clustering, one may ask whether the results will be as accurate.  We investigate and address this question from the perspective of how tiny subsamples of observations and separately features affect clustering results.  First, note that if a tiny fraction of observations is subsampled, then by chance some of the clusters may not be represented; this is especially the case for large $K$ or for uneven cluster sizes.  Existing consensus clustering approaches typically apply a clustering algorithm with fixed $K$ to each subsample, but this practice would prove detrimental for our approach.  Instead, we propose to choose the number of clusters on each minipatch adaptively.  While there are many techniques in the literature to do so that could be employed with our method \citep{fred2005combining, fred2002evidence}, we are motivated to choose the number of clusters very quickly with nearly no additional computation.  Hence, we propose to exclusively use hierarchical clustering on each minipatch and to cut the tree at the $h$ quantile (typically set to 0.95) of the dendrogram height to determine the number of clusters and cluster membership.  This approach is not only fast but adaptive to the number of clusters present in the minipatch, and results change smoothly with cuts at different heights.  Our empirical results reveal that this approach performs well on minipatches and we specifically investigate its utility, sensitivity, and tuning of $h$ in Appendix~F; importantly, we find that setting $h = .95$ to nearly universally yields the best results and hence we suggest fixing this value.  Additionally, we provide details on hyper-parameters, tuning, and stopping criteria in Appendix~F.

Next, one may ask how subsampling the features in minipatches affects clustering accuracy.  Obviously for high-dimensional data in which only a small number of features are relevant for differentiating clusters, subsampling minipatches containing the correct features would improve results.  We address such possibilities in the next section.  But if this is not the case, would clustering accuracy suffer?  Since we apply hierarchical clustering which takes distances as input, we seek to understand how far off our distance input can be when we employ sub-samples of features. To this end, we consider distances that can be written in the form of the sum; this includes popular distances like the Manhattan or squared Euclidean distance, among others.  The following result probabilistically bounds the deviations of the distances computed using only a subset of features:

\begin{prop}\label{pr:hoef} For $\epsilon>0$ and $|\hat{d}_{i,i'} - d^*_{i,i'}|\in[0,1]$, $P\left(|\hat{d}_{i,i'} - d^*_{i,i'}|\ge \epsilon\right)  \le 2\exp \left(\frac{-2m\epsilon^2}{(1-\frac{m-1}{M})}\right)$,

\end{prop}
where $d_{i,i'}^* = \frac{1}{M}f(x_i,x_{i'})$ is the distance between observations $x_i$ and $x_{i'}$ using the full set of $M$ features, and $\hat{d}_{i,i'} = \frac{1}{m}\sum_{j\in {j_1,...,j_m}}f(x_{i,j},x_{i',j})$ is the distance using a subset of $m$ features. This is derived from the Hoeffding inequality \citep{10.1214/aos/1176342611}.  This result states that the probability that distances computed on minipatches are far off from original distances is small, under the worst-case scenario. This provides some reassurances that clustering accuracy based on subsampling features should not greatly suffer.  While smaller minipatches yield faster computations, there may be a slight trade-off in terms of clustering accuracy. Our empirical results in Appendix~F suggest that such a trade-off is generally slight or negligible, so we can typically utilize smaller minipatches.

\begin{algorithm*}\label{algo:rand}~
\SetAlgoLined
\noindent{\textbf{Input}}: $\boldsymbol{X}$, $n$, $m$, $\boldsymbol{\mathcal{V}}^{(0)}$, $\boldsymbol{\mathcal{D}}^{(0)}$, $h$; 

 \While {stopping criteria not meet} {
\begin{minipage}{.9\textwidth}
 \begin{enumerate}
    \item Obtain minipatch $\boldsymbol{X}_{I_t,F_t}$ $\in \mathbb{R}^{n \times m}$ by subsampling $n$ observations $I_t\subset \{1,...,N\}$ and $m$ featrues $F_t\subset \{1,...,M\}$, without replacement; 
    
    \begin{itemize}
        \item \textit{MPCC subsamples uniformly at random;}
        \item \textit{MPACC uses the adaptive observation sampling scheme only;}     \item \textit{IMPACC uses both adaptive feature and observation sampling schemes simultaneously;} 
    \end{itemize}
        
        \item Obtain estimated clustering result $\mathcal{C}^{(t)}$ by fitting hierarchical clustering to $\boldsymbol{X}_{I_t,F_t}$ and cut tree at $h$ height quantile;
        
        \item Update co-clustering membership matrix
        $\boldsymbol{\mathcal{V}}$ and co-sampling matrix $\boldsymbol{\mathcal{D}}$:
    $\boldsymbol{\mathcal{V}}^{(t)}(i,i') = \boldsymbol{\mathcal{V}}^{(t-1)}(i,i') + \mathbb{I}(\mathcal{C}^t_i = \mathcal{C}^t_{i'}); \quad \boldsymbol{\mathcal{D}}^{(t)}(i,i') = \boldsymbol{\mathcal{D}}^{(t-1)}(i,i') + \mathbb{I} (i\in  I_t, i' \in I_t))$;
 \end{enumerate}
  \end{minipage}
}
 Calculate consensus matrix $\boldsymbol{\mathcal{S}}(i,i') = \frac{ \boldsymbol{\mathcal{V}}^{(T)}(i,i')}{ \max(1,\boldsymbol{\mathcal{D}}^{(T)}(i,i'))}$;

Obtain final clustering result $\hat\Pi$ by using $\boldsymbol{\mathcal{S}}$ as a similarity matrix;

 \textbf{Output}: $\boldsymbol{\mathcal{S}}$, $\hat\Pi$.\
 \caption{Minipatch Consensus Clustering}
\end{algorithm*}

We have introduced minipatch consensus clustering (MPCC) using random subsamples of both features and observations.  The advantage of this approach is its computational speed, which our empirical results in Section~\ref{sec:result} suggest is on the order of standard clustering approaches such as hierarchical and spectral clustering (hence confirming Proposition~\ref{le:tc}).  But, one may ask whether clustering results can be improved by perhaps optimally sampling observations and/or features instead of using random sampling.  Some have suggested such possibilities in the context of consensus clustering \citep{duarte2012adaptive, parvin2013data,topchy2004adaptive,ren2017weighted}; we explore it and develop new approaches for this in the following sections.

\subsection{Minipatch Adaptive Consensus Clustering (MPACC)}\label{sec:MPACC}

One may ask whether it is possible to improve upon minipatch consensus clustering in terms of both speed and clustering accuracy by adaptively sampling observations.  For example, we may want to sample observations that are not well clustered more frequently to learn their cluster assignments faster. In the method MiniPatch Adaptive Consensus Clustering (MPACC), we propose to dynamically update sampling weights, with a focus on observations that are difficult to be clustered and that are less frequently sampled. In addition, we leverage the adaptive weights by designing a novel observation sampling scheme.   
\begin{algorithm}[h]\label{algo:weight_i}
\SetAlgoLined

  \noindent{\textbf{Input}}: $\boldsymbol{\mathcal{S}}^{(t-1)}$, $\{I_{l}\}_{l=1}^{t-1}$,  $\alpha_I$;
    \begin{enumerate}
        \item Calculate sample uncertainty $\boldsymbol{u}_i =   \frac{1}{N}\sum_{i'=1}^N \boldsymbol{\mathcal{S}}^{(t-1)}_{i,i'}(1-\boldsymbol{\mathcal{S}}^{({t-1})}_{i,i'}) \times \frac{t-1}{\sum_{l=1}^{t-1} \mathbb{I}(i\in I_l)}$;

        \item Update observation weight vector $\boldsymbol{w_I}^{(t)} = \alpha_I \boldsymbol{w_I}^{(t-1)} + (1-\alpha_I) \frac{\boldsymbol{u}}{\sum_{i=1}^N \boldsymbol{u}_i}$;
 
    \end{enumerate}

 \textbf{Output}: $\boldsymbol{w_I}^{(t)}$.
 \caption{Weight updating in adaptive observation sampling scheme}
\end{algorithm}
Specifically, we propose to dynamically update observations weights by adjusted confusion values. To measure the level of clustering uncertainty, confusion values are derived from consensus matrix, given by $confusion_i = \frac{1}{N}\sum_{i'=1}^N\boldsymbol{\mathcal{S}}_{i,i'}(1-\boldsymbol{\mathcal{S}})_{i,i'}$ for observation $i$. A larger confusion value near $0.25$ indicates poorer clustering with unstable assignments, and the minimum confusion value $0$ suggests perfect clustering. Note that confusions tend to grow with iterations because more consensus values are updated from the initial value $0$. Therefore, a large confusion value due to oversampling cannot truly reflect the level of uncertainty. To eliminate bias caused by oversampling and to upweight less frequently sampled observations, we further adjust confusion values by sampling frequencies of observations in previous iterations, as presented in Algorithm~\ref{algo:weight_i}.

The next question is, how do we leverage the weights to dynamically construct minipatches as the number of iterations grows? One simple solution is to probabilistically subsample with probability ($Prob$) proportional to the weights. But the problem with this approach is that the clustering performance will be compromised if we only tend to sample uncertain and difficult observations.  To resolve such drawback, we develop an exploitation and exploration plus probabilistic ($EE+Prob$) sampling scheme (Algorithm~\ref{algo:adj_feature_ep}).  The scheme consists of two sampling stages: a burn-in stage and an adaptive stage. The purpose of the burn-in stage is to explore the entire observation space and ensure every observation is sampled several times.  During the next adaptive stage, observations with the levels of uncertainty greater than a threshold are classified into the high uncertainty set, and the algorithm exploits this set by sampling $\gamma^{(t)}$ proportion of observations using probabilistic sampling.  Here, $\gamma^{(t)}\in[0.5,1]$ is a monotonically increasing parameter that controls sampling size in the exploitation and exploration step. Meanwhile, the algorithm explores the rest of observations with uniform weights, to avoid exclusively focusing on difficult observations. The reason why we randomly sample the observations that we are confident about is that, we need to include a fair amount of easy-to-cluster observations to construct well-defined clusters in each minipatch so as to better cluster the uncertain ones.  We also propose to use the $EE+Prob$ scheme as our adaptive feature sampling scheme, which is discussed in Section~\ref{sec:IMPACC}.

\begin{algorithm}
\SetAlgoLined
 
\noindent{\bf Input}:$t$, $n$, $N$, $E$, $\{\gamma^{(t)}\}$, $\boldsymbol{w_I}^{(t-1)}$, $\{\tau\}$;

\noindent{\bf Initialization}:$Q = [\frac{N}{n}]$, $\mathcal{I} = \{1,...,N\}$;\

  \eIf{$t \le E\cdot Q$}{  \tcp*[h]{Burn-in stage} 
   
    \eIf{$mod_Q(t)=1$}{  \tcp*[h]{  New epoch}
    
      Randomly reshuffle feature index set $\mathcal{I}$ and partition into disjoint sets  $\{\mathcal{I}_q\}_{q=0}^{Q-1}$;\
      
   }{ 
   
   Set $I_t = \mathcal{I}_{mod_Q(t)}$;\
  }
  
   }{{\tcp*[h]{Adaptive stage}}

  \begin{enumerate}
      \begin{minipage}{1\textwidth}

        \item Update observation weights $\boldsymbol{w_I}^{(t)}$ by Algorithm~\ref{algo:weight_i};
 \item  Create high uncertainty set $\mathcal{H}_I = \{i \in \{1,...,N\}: \boldsymbol{w_I}^{(t)}_i> {\tau_{\boldsymbol{w_I}^{(t)}}}\}$;
     \item Exploitation: sample $\min(n,\gamma^{(t)}|\mathcal{H}_I|)$ observations $I_{t,1} \subseteq \mathcal{H}_I$ with probability $\boldsymbol{w_I}_{\mathcal{H}_I^t}$;
                
        \item Exploration: sample $(n-\min(n,\gamma^{(t)}|\mathcal{H}|_I))$ observations $I_{t,2} \subseteq \{1,...,N\}\backslash\mathcal{H}_I$ uniformly at random; 
        
        \item Set  $I_t = I_{t,1}\cup I_{t,2}$;
  
   \end{minipage}

   \end{enumerate}
 }
   
 \textbf{Output}: $I_t$.
 
 \caption{Adaptive Observation (Features) Sampling Scheme - $EE+Prob$}\label{algo:adj_feature_ep}~
\end{algorithm}

\subsubsection{Relation to Existing Literature}

Several have suggested similar weight updating approaches in the consensus clustering literature. \citet{ren2017weighted} also obtain observation weights by confusion values as in our method. The difference is that, their methods only use the weight scheme at the final clustering step rather than adaptive sampling. On the other hand, similar to our adaptive weight updating scheme, \citet{duarte2012adaptive,topchy2004adaptive,parvin2013data} iteratively update weights depending on clustering history. However, these existing methods utilize probabilistic sampling, so they would largely suffer from biased sampling and inaccurate results by only focusing on hard observations. However, instead of probabilistic sampling, we design the $EE+Prob$ sampling scheme to leverage the weights, which is inspired by the exploration and exploitation ($EE$) scheme from multi-arm bandits \citep{bouneffouf2019survey,slivkins2019introduction} and also employed for feature selection with minipatches in \citet{yao2020feature}. Compared to the latter, the innovation in our approach is to combine the advantages of probabilistic sampling and exploitation-exploration sampling which proves to have particular advantages for clustering. Comparisons with other possible sampling schemes proposed in the literature are shown in Appendix~F.

\subsection{Interpretable Minipatch Adaptive Consensus Clustering (IMPACC)}\label{sec:IMPACC} 

One major drawback of consensus clustering is that it lacks interpretability into important features.  This is especially important for high-dimensional data like that in bioinformatics where we expect only a small subset of features to be relevant for determining clusters.  To address this, we develop a novel adaptive feature sampling approach termed Interpretable Minipatch Adaptive Consensus Clustering (IMPACC) that learns important features for clustering and hence improves clustering accuracy for high-dimensional data.  In clustering, two types of approaches to determine important features have been proposed.  One is to obtain a sparse solution by solving an optimization problem \citep{witten2010framework, wang2018sparse,wang2021integrative}, and another one is to rank features by their influence to results \citep{yu2019ensemble, dash2000feature, zhao2007spectral}.   However, in data sets with a large number of observations and features, both kinds of methods suffer from significant computational inefficiency.  So the question we are interested in is, can we achieve fast, accurate and reliable feature selection within the consensus clustering process with minipatches? We address this question by proposing a novel adaptive feature weighting method that measures the feature importance in each minipatch and then ensembles the results to increase the weights of the important features. Given these adaptive feature weights, we can then utilize our adaptive sampling scheme proposed in Algorithm~\ref{algo:adj_feature_ep} to more frequently sample important features.

Outlined in Algorithm~\ref{algo:weight_f}, we propose an adaptive feature weighting scheme by testing whether each feature is associated with the estimated cluster labels on that minipatch.  To do so, we use a simple ANOVA test in part, because it is computationally fast and only requires one matrix multiplication.  Based on the p-values from these tests, we establish an important feature set, $A$, and obtain the importance scores as the frequency of features being classified into this feature set over iterations. Then the feature sampling weights are dynamically updated with learning rate $\alpha_F$.  Therefore, by ensembling feature importance obtained from each iteration, we are able to simultaneously improve clustering accuracy and build model interpretability from resulting feature weights, with minimal sacrifices in terms of computational time.

\begin{algorithm}\label{algo:weight_f}~
\SetAlgoLined

\noindent{\textbf{Input}}: $\boldsymbol{X}_{I_t,F_t}$, $\mathcal{C}^{(t-1)}$, $\{F_{l}\}_{l=1}^{t-1}$, $\{A^l\}_{l=1}^{t-2}$, $\{\eta\}$, $\boldsymbol{w_F}^{(t-1)}$, $\alpha_F$;
   \begin{enumerate}
      \begin{minipage}{1\textwidth}

    \item For each feature $j \in F_{t-1}$, conduct ANOVA test between features $j$ and $\mathcal{C}^{(t-1)}$, record p-value $p_j^{(t-1)}$;
        
    \item Create a feature support $A^{(t-1)} \subseteq F_{t-1}$: $A^{(t-1)} =\{j\in\{1,...,m\}: p^{(t-1)}_j <\eta_{p^{(t-1)}}\}$;
       
       \item Update feature weight vector $\boldsymbol{w_F}^{(t+1)} \in \mathbb{R}^M$ by ensembling feature supports $\{A^l\}_{l=1}^{(t-1)}$:
       
       $$
       \boldsymbol{w_F}_j^{(t)} =\alpha_F  \boldsymbol{w_F}_j^{(t-1)} + (1-\alpha_F)\frac{\sum\limits_{l=1}^{t-1}\mathbb{I}(j\in F_l,j\in {A}^l)}{\max(1,{\sum\limits_{l=1}^{t-1}}\mathbb{I} (j\in  F_l))};
       $$
\end{minipage}
   \end{enumerate}
 \textbf{Output}: $\boldsymbol{w_F}^{(t)}$.
 \caption{Weight updating in adaptive feature sampling scheme}
\end{algorithm}

We propose to utilize the same type of $EE+Prob$ sampling scheme (Algorithm~\ref{algo:adj_feature_ep}) given our feature weights to learn the important features for clustering. Such a scheme exploits the important features and samples these more frequently as the algorithm progress.   But it also balances exploring other features to ensure that potentially important features are not missed.  Our final IMPACC algorithm then utilizes both adaptive observation sampling and adaptive feature sampling to both improve computation and clustering accuracy while also providing feature interpretability.  Utilizing minipatches in consensus clustering allows us to develop these innovative adaptive sampling schemes and be the first to propose feature learning in this context.

Even though IMPACC has several hyper-parameters, in practice, our methods are quite robust and reliable to parameter selections, and generally give a strong performance under default parameter settings. Therefore, we are freed from the computationally expensive hyper-parameter tuning process and its computational burdens. We include a study on learning accuracy with different levels of hyper-parameters, the default values, and also suggest a data-driven tuning process in Appendix~F.

Overall, the proposed MPACC with only adaptive sampling on observation is more suitable for data of no or little sparsity; and IMPACC, which adaptively subsamples both observations and features in minipatch learning, can be more useful when dealing with high dimensional and sparse data set in bioinformatics. It enhances model accuracy, scalability, and interpretability, by focusing on uncertain observations and important features in an efficient manner. Our empirical study in Section~\ref{sec:result} demonstrates the major advantages of the IMPACC method in terms of clustering quality, feature selection accuracy and computation saving.

\section{Empirical Studies}\label{sec:result} 
In this section we assess the performance of IMPACC and MPCC with application to a high dimensional and high noise synthetic simulation study in Section~\ref{sec:simu} and four large-scale real data sets in Section~\ref{sec:real} , in comparison with several conventional clustering strategies.

\subsection{Synthetic Data}{\label{sec:simu}}

We evaluate the performance of MPCC and IMPACC in terms of clustering accuracy and computation time with widely used competitors, and compare IMPACC's feature selection accuracy with the existing sparse feature selection techniques. Simulations are conducted under three scenarios: sparse, weak sparse and no sparse. We only show the results of sparse simulation, as it is the best representative of high dimensional bioinformatics data, and results of the other two scenarios can be found in Appendix~C. 

In the sparse simulation study, each data set is created from a mixture of Gaussian with block-diagonal covariance matrix $\Sigma = \textbf{I}_{\frac{M}{5},\frac{M}{5}} \otimes (\rho\cdot \textbf{1}_5\cdot\textbf{1}_5^T+(1-\rho)\cdot\textbf{I}_{5,5})$, where $\otimes$ denotes the Kronecker product. The parameter $\rho$ is set to be $0.5$. We set the number of observations, features and clusters to be $N = 500$, $M = 5,000$, $K = 4$, respectively, and the numbers of observations in each cluster are $20$, $80$, $120$, $280$. The means of features in synthetic data is $\mu = [\mu_{k},\mu_{0}]$, where $\mu_{k} \in \mathbb{R}^{25}$ and $\mu_{0} = \textbf{0}_{4975}$ are the means of $25$ signal features and $4,975$ noise features, respectively. The signal-to-noise (SNR) ratio is defined as the L2-norm of feature means: $SNR = \|\mu\|_2$. In order to assess feature selection capability, synthetic data is generated with $SNR$ ranging from 1 to 8. Specifically, the signal features are generated with $\mu_1 = \frac{SNR}{5}\cdot\textbf{1}_{25}$, $\mu_2 = (\frac{SNR}{5}\cdot\textbf{1}_{13}^T, -\frac{SNR}{5}\cdot\textbf{1}_{12}^T)^T$, $\mu_3 = (-\frac{SNR}{5}\cdot\textbf{1}_{13}^T, \frac{SNR}{5}\cdot\textbf{1}_{12}^T)^T$, $\mu_4 = -\frac{SNR}{5}\cdot\textbf{1}_{25}$. Data with higher SNR ratio has more informative signal features so is easier to be clustered.  For all clustering algorithms, we assume oracle number of clusters $K$. Hierarchical clustering is applied as the final algorithm in IMPACC and MPCC, with number of iterations determined by an early stopping criteria, as described in Appendix~B. And we have exactly the same setting as those of MPCC in regular consensus clustering, including the number of iterations. Ward’s minimum variance method with Manhattan distance is used in all hierarchical clustering related methods.

\begin{figure*}[!h]
\centering\includegraphics[scale=0.08]{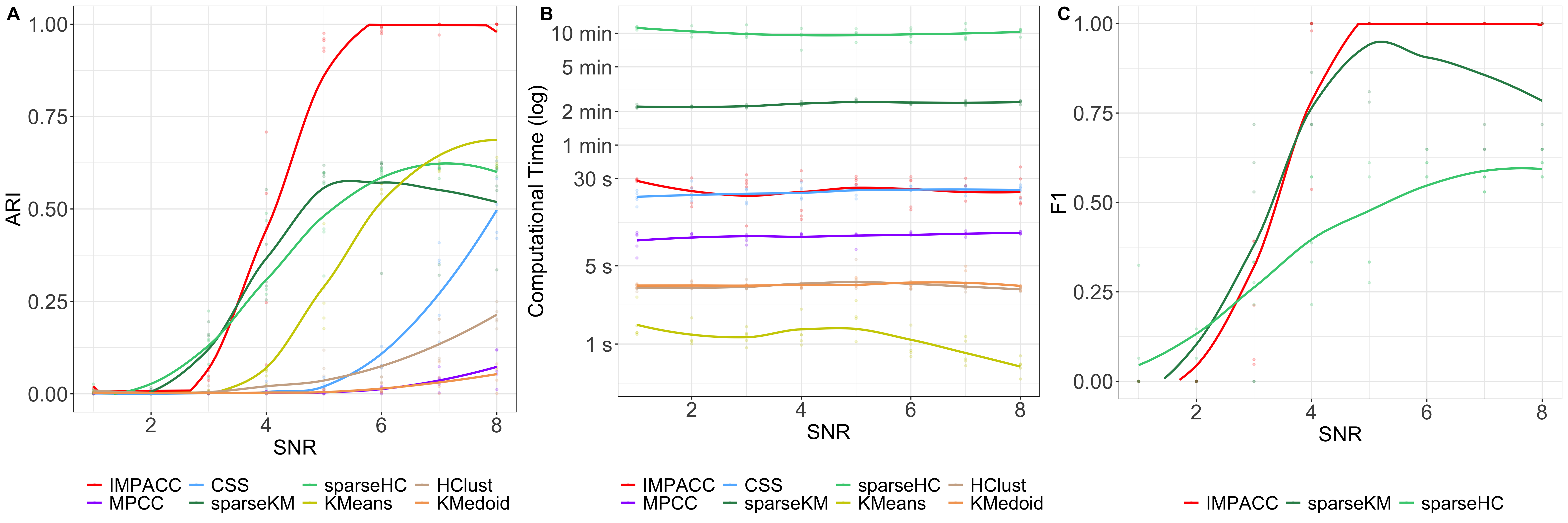}
\caption{Clustering performance (ARI), feature selection accuracy (F1 score), and computation time on sparse synthetic data sets. (A) ARI (higher is better) of estimated grouping; (B) computation time in log seconds; (C) F1 score for signal feature estimates. IMPACC has superior performance over competing methods in clustering and feature selection accuracy with significant computation saving.}\label{fig:simu}
\end{figure*}

We use adjusted rand index (ARI) to evaluate the clustering performance, and F1 score to measure feature selection accuracy, which both range in $[0,1]$, with a higher value indicating higher accuracy. The averaged results over 10 repetitions are shown in Figure~\ref{fig:simu}. Overall, IMPACC yields the best clustering performance over all competing methods with the highest ARI in most of the $SNR$ settings. Comparing feature selection performance, IMPACC has perfect recovery on informative features, with an F1 score equaling to $1$ when $SNR$ is large, and is significantly better than sparseKM and sparseHC. Additionally, IMPACC achieves significantly major computational advantages comparing to sparse feature selection clustering strategies. All of the computation time is recorded on a laptop with 16GB of RAM (2133 MHz) and a dual-core processor (3.1 GHz). Note that we only show results of the sparse simulation scenario in Figure~\ref{fig:simu}, and we include the other two scenarios in Appendix~C. Our methods are still dominant in noisy and weak sparse situations, but IMAPCC shows little improvement on the no-sparsity scenario when all the features are relevant.

\subsection{Case Studies on Real Data}\label{sec:real}
We apply our methods to three RNA-seq data sets and one image data set with known cluster labels, whose information is reported in Table~\ref{table:dataVal}. In the RNA-seq data, gene expressions are transformed by $x\mapsto\log_2(1+x)$ before conducting clustering algorithms; the image data set is adjusted to be within the range $[0,1]$. With the same settings in Section~\ref{sec:simu}, we evaluate the learning performance of MPCC and IMPACC with existing methods, with the number of clusters being oracle.   

\begin{table}
\centering
\begin{tabular}
 {ccccc}
 & \bf PANCAN & \bf Brain & \bf Neoplastic & \bf COIL20 \\
\toprule
\bf Data type & RNA-seq   & scRNA-seq & scRNA-seq & Image\\
\hline
\bf Tissue & tumor cells  & brain cells& neoplastic  infiltrating cells & \\
\hline
\bf \# clusters & 5 & 4 & 7 & 20\\
  
\hline
\bf \# observations &  761&  366  & 3,576 & 1,440\\
\hline
\bf \# features &  13,244& 21,413 & 28,805 & 1,024\\
\hline
\bf \% zeros & 14.2\% & 80.06\% & 81.36\% & 34.38\%\\
\hline
\bf citation & \citet{weinstein2013cancer} & \citet{darmanis2015survey} & \citet{darmanis2017single} & \citet{Nene96columbiaobject}\\
\hline
\bf GEO accession code & GSE43770 & GSE67835 &  GSE84465 &\\
\bottomrule
\end{tabular}\caption{Data sets used in empirical study.}\label{table:dataVal}
\end{table}

\begin{figure*}[!h]
\centering\includegraphics[scale=0.1]{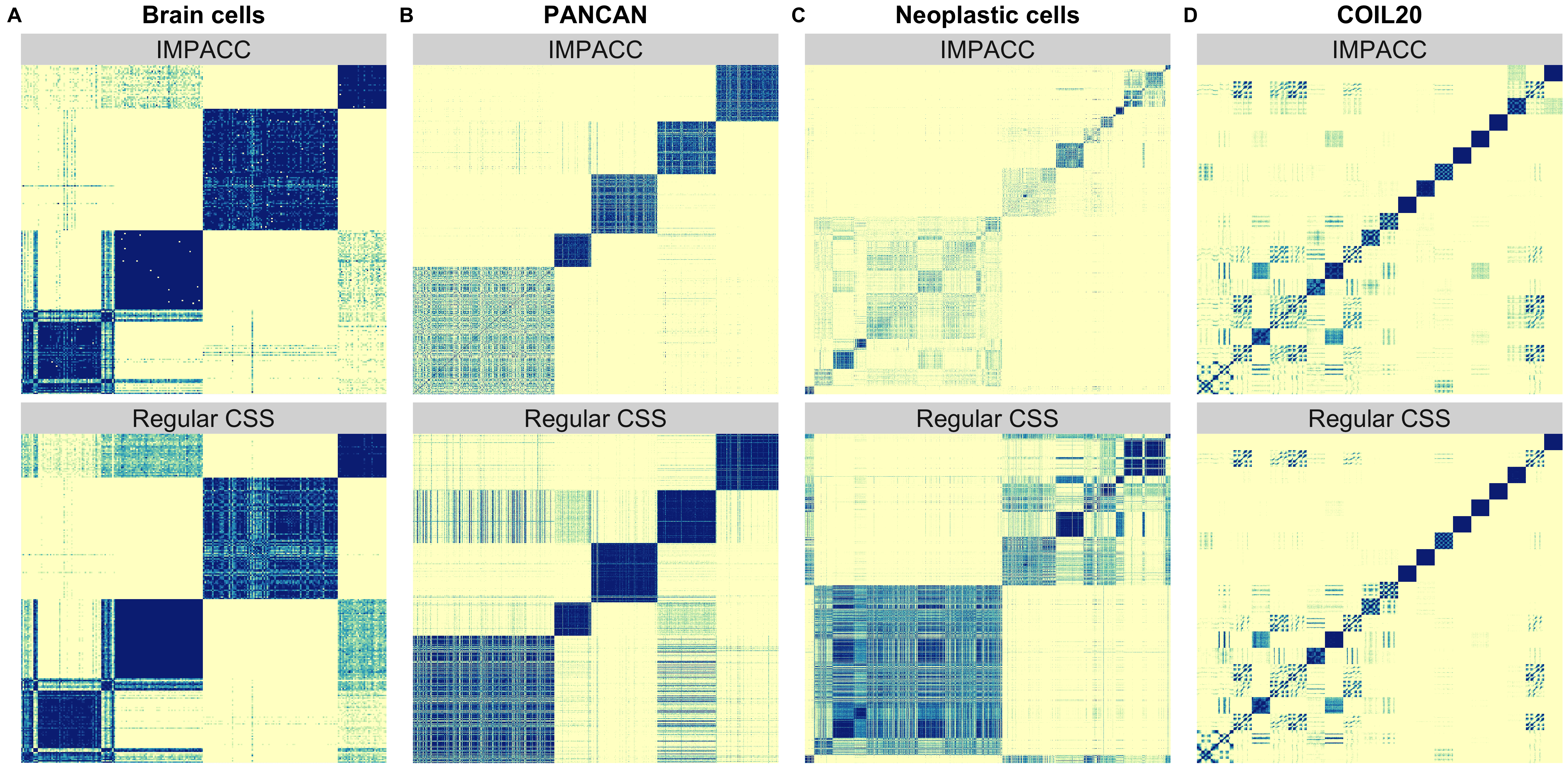}
\caption{Heatmaps of final consensus matrix derived from IMPACC and  consensus clustering respectively, using oracle number of clusters. Darker color indicates higher consensus value. }\label{fig:css}
\end{figure*}

Table~\ref{table:real} summarizes clustering results on real data sets. IMPACC consistently outperforms all competing methods at discovering known clusters with the highest ARI score, and it demonstrates major computational advantages, sometimes even beating hierarchical clustering. Clustering followed by dimension reduction via tSNE can have faster and better clustering accuracy for some of the data sets, but they fail to provide interpretability in terms of feature importance. Even though single cell RNA-seq specific methods Seurat and CS3 have comparable accuracy in the brain data set, these methods select genes with high variance before performing clustering algorithm and do not provide inherent interpretations of important genes. Note that {\tt R} failed to apply sparseHC to large genomics data due to excessive demand on computing memory. Further, even though MPCC has slightly lower ARI than IMPACC, it still yields better performance in learning accuracy over consensus and standard methods, and it achieves the fastest computational speed over all other methods in most of the data sets, excluding K-Means clustering. Additionally, we visualize the consensus matrices of IMPACC and compare to that of regular consensus clustering in Figure~\ref{fig:css}. We can conclude that IMPACC is able to produce more accurate consensus matrices, with clearer diagonal blocks of clusters and less noise on off-diagonal entries. 

\begin{table*}
\centering
\begin{tabular}{lrrrrrrrr}
\toprule
\multicolumn{1}{c}{ } & \multicolumn{4}{c}{\bf ARI} & \multicolumn{4}{c}{\bf Time (s)} \\
\cmidrule(l{3pt}r{3pt}){2-5} \cmidrule(l{3pt}r{3pt}){6-9}

  & PANCAN & brain  & neoplastic & COIL20 & PANCAN & brain & neoplastic& COIL20\\
\midrule
IMPACC (HC) &  0.939 & \bf 0.978 & \bf 0.908 & 0.744 & 18.491 & 29.204 & 1843.033 & 115.119 \\
IMPACC (Spec) & 0.828 & 0.924 & 0.856 & 0.711 \\
\hline
MPCC (HC) & 0.922 & 0.844 & 0.808 & 0.715 & 13.021& 9.631& 2170.650& 91.319\\
MPCC (Spec) & 0.833 & 0.804 & 0.842& 0.67 \\
\hline
Consensus (HC) & 0.761 & 0.610 & 0.398& 0.719 & 88.380 & 30.787 & 8866.810& 1730.019\\
Consensus (Spec) & 0.770 & 0.548 & 0.534 & 0.665\\
\hline
sparseKM & 0.784 & 0.961 & 0.486 & 0.459& 1236.009 & 375.658 & 62580.775& 207.68\\
\hline
sparseHC & N/A & 0.247 & N/A & 0.158& N/A & 1540.236& N/A& 1867.174\\
\hline
KMeans & 0.797 & 0.588 & 0.513 & 0.542& 2.154 & 1.558 & 97.083& 0.305\\
\hline
KMedoid & 0.795 & 0.255 & 0.160 & 0.532& 30.494 & 12.261 & 6063.345& 9.079\\

\hline
HClust & 0.769 & 0.613 & 0.413 & 0.688& 28.016 & 13.835 & 6052.748& 6.274\\
\hline
Spectral & 0.776 & 0.575 & 0.671 & 0.561& 19.640 & 7.561 & 1463.087& 69.546\\
\hline

tSNE+KMeans   &0.863& 0.702   & 0.346 &0.614&  29.222 &  15.008  & 1577.265 & 8.859    \\
\hline
tSNE+KMedoid &0.960& 0.944 & 0.342 &  0.740 & 29.390 & 15.023  & 1582.872 & 13.138\\
\hline

tSNE+HClust &\textbf{0.990}&  0.950 & 0.445 &  \textbf{0.760}& 29.240  & 15.001  & 1577.719 & 8.906 \\
\hline
tSNE+Spectral &0.789& 0.726 & 0.725 & 0.641 & 44.824  & 17.541  & 3126.325    & 121.169  \\
\hline

Seurat && 0.908 & 0.683 &  &  & 4.285 & 25.019& \\
\hline
CS3 && 0.978 & 0.453 &  &  & 134.920 & 1486.074& \\
\bottomrule

\end{tabular}
\caption{Clustering performance (ARI) and computation time in seconds on real data sets with known cluster labels. The IMPACC method outperforms others in clustering performance, with significant improvements on computational cost compared to sparseKM, sparseHC and consensus clustering. The MPCC method also yields comparable clustering performance, and achieves the fastest computational speed excluding K-Means clustering.}
\label{table:real}
\end{table*}

IMPACC further provides interpretablility in terms of feature importance. 19 of top 25 genes with high importance (feature score $\in[0.78,1]$) in brain cell data set are enriched/enhanced in brain; in top 25 genes (feature score $\in[0.933,1]$) in the PANCAN tumor data set, 11 genes are prognostic cancer markers and 13 are enriched/enhanced in tissues; 9 of the top 25 genes (feature score $\in[0.52,1]) $ in neoplastic cells are enriched in the brain, and 17 genes are prognostic cancer markers. For example, BCAN is highly relevant to tumor cell migration with contribution to nervous system development, and OPALIN is a known marker in oligodendrocytes \citep{darmanis2017single}. The gene information is sourced from the Human Protein Atlas \citep{ponten2008human}, and more details on significant genes can be found in Appendix~D.

To further evaluate the model interpretability of IMPACC, sparseKM and sparseHC, we perform pathway analysis on the most important genes discovered by each method. We determined genes as important if their feature importance scores were higher than the mean plus one standard deviation of all scores.  IMPACC is able to identify a larger set of important genes with more discrepancy between signal and noise genes. By performing KEGG pathway analysis, we find the important genes obtained from IMPACC are enriched in much more biological meaningful pathways with smaller p-values, comparing to those identified by sparseKM and sparseHC. For example, as shown in Table ~\ref{table:path}, $578$, $89$ and $27$ important genes are detected by IMPACC, sparseKM and sparseHC, respectively in the brain cell data. The top enriched KEGG pathways from IMPACC is GABAergic synapse, which is the main neurotransmitter in adult mammalian brain \citep{watanabe2002gaba}. Therefore, IMPACC provides accurate and reliable interpretations on scientifically important genes. Additional pathway analyses are detailed in Appendix~D . 

\begin{table*}
\centering
\resizebox{\textwidth}{!}{\begin{tabular}{lrrrrrrrrr}
\toprule
\multicolumn{1}{c}{ } & \multicolumn{3}{c}{\bf IMAPCC} & \multicolumn{3}{c}{\bf sparseKM }& \multicolumn{3}{c}{\bf sparseHC } \\
\cmidrule(l{3pt}r{3pt}){2-4} \cmidrule(l{3pt}r{3pt}){5-7} \cmidrule(l{3pt}r{3pt}){8-10} 
  & Pathway & Name  & p-value & Pathway & Name  & p-value& Pathway & Name  & p-value\\
\midrule
1 & hsa04727 & GABAergic synapse & 2.357e-15 & hsa04964 & Proximal tubule bicarbonate reclamation & 1.3191e-07 & hsa04976 & Bile secretion & 0.0011\\
2 & hsa04911 & Insulin secretion & 1.457e-12 & hsa04727 & GABAergic synapse & 8.602e-06 & hsa04964 & Proximal tubule bicarbonate reclamation & 0.0013\\
3 & hsa04721 & Synaptic vesicle cycle & 2.208e-12 & hsa04976 & Bile secretion & 0.0001 & hsa04724 & Glutamatergic synapse & 0.0022\\
4 & hsa04978 & Mineral absorption & 3.492e-09 & hsa04978 & Mineral absorption & 0.0003 & hsa04919 & Thyroid hormone signaling pathway & 0.0027\\
5 & hsa04971 & Gastric acid secretion & 1.121e-08 & hsa04919 & Thyroid hormone signaling pathway & 0.00052 & hsa01230 & Biosynthesis of amino acids & 0.0131\\
\bottomrule

\end{tabular}}
\caption{Top 5 enriched KEGG pathways in brain cell data set}
\label{table:path}
\end{table*}

\section{Discussion}\label{sec:dis}
We have proposed novel and powerful methodologies for consensus clustering using minipatch learning with random or adaptive sampling schemes. We have demonstrated that both MPCC and IMPACC are stable, robust, and offer superior performance than competing methods in terms of computational accuracy. Further, the approaches offer dramatic computational savings with runtime comparable to hierarchical or spectral clustering.   Finally, IMPACC offers interpretable results by discovering features that differentiate clusters.  This method is particularly applicable to sparse, high-dimensional data sets common in bioinformatics.  Our empirical results suggest that our method might prove particularly important for discovering cell types from single-cell RNA sequencing data. Note that while our methods offer computational advantages over consensus clustering for all settings, our method does not seem to offer any dramatic improvement in clustering accuracy for non-sparse and non-high-dimensional data sets.  In future work, one can further optimize computations through memory-efficient management of the large consensus matrix and through hashing or other approximate schemes.  Overall, we expect IMPACC to become a critical instrument for clustering analyses of complicated and massive data sets in bioinformatics as well as a variety of other fields.

\section*{Acknowledgements}
The authors would like to thank Zhandong Liu and Ying-Wooi Wan for helpful discussions on single-cell sequencing as well as Tianyi Yao for helpful discussions on minipatch learning. 

\section*{Funding}
This work has been supported by NSF DMS-1554821 and NIH 1R01GM140468.

\textit{Conflict of Interest: none declared.
}

\bibliographystyle{plainnat}
\bibliography{ref.bib}

\begin{thebibliography}{54}
\providecommand{\natexlab}[1]{#1}
\providecommand{\url}[1]{\texttt{#1}}
\expandafter\ifx\csname urlstyle\endcsname\relax
  \providecommand{\doi}[1]{doi: #1}\else
  \providecommand{\doi}{doi: \begingroup \urlstyle{rm}\Url}\fi

\bibitem[Analoui and Sadighian(2007)]{analoui2007solving}
Morteza Analoui and Niloufar Sadighian.
\newblock Solving cluster ensemble problems by correlation’s matrix \& ga.
\newblock In \emph{Intelligent Information Processing III: IFIP TC12
  International Conference on Intelligent Information Processing (IIP 2006),
  September 20--23, Adelaide, Australia 3}, pages 227--231. Springer, 2007.

\bibitem[Azimi et~al.(2006)Azimi, Mohammadi, Analoui,
  et~al.]{azimi2006clustering}
Javad Azimi, Mehdi Mohammadi, Morteza Analoui, et~al.
\newblock Clustering ensembles using genetic algorithm.
\newblock In \emph{2006 International Workshop on Computer Architecture for
  Machine Perception and Sensing}, pages 119--123. IEEE, 2006.

\bibitem[Azimi et~al.(2007)Azimi, Abdoos, and Analoui]{azimi2007new}
Javad Azimi, Monireh Abdoos, and Morteza Analoui.
\newblock A new efficient approach in clustering ensembles.
\newblock In \emph{International Conference on Intelligent Data Engineering and
  Automated Learning}, pages 395--405. Springer, 2007.

\bibitem[Bouneffouf and Rish(2019)]{bouneffouf2019survey}
Djallel Bouneffouf and Irina Rish.
\newblock A survey on practical applications of multi-armed and contextual
  bandits.
\newblock \emph{arXiv preprint arXiv:1904.10040}, 2019.

\bibitem[Darmanis et~al.(2015)Darmanis, Sloan, Zhang, Enge, Caneda, Shuer,
  Gephart, Barres, and Quake]{darmanis2015survey}
Spyros Darmanis, Steven~A Sloan, Ye~Zhang, Martin Enge, Christine Caneda,
  Lawrence~M Shuer, Melanie G~Hayden Gephart, Ben~A Barres, and Stephen~R
  Quake.
\newblock A survey of human brain transcriptome diversity at the single cell
  level.
\newblock \emph{Proceedings of the National Academy of Sciences}, 112\penalty0
  (23):\penalty0 7285--7290, 2015.

\bibitem[Darmanis et~al.(2017)Darmanis, Sloan, Croote, Mignardi, Chernikova,
  Samghababi, Zhang, Neff, Kowarsky, Caneda, et~al.]{darmanis2017single}
Spyros Darmanis, Steven~A Sloan, Derek Croote, Marco Mignardi, Sophia
  Chernikova, Peyman Samghababi, Ye~Zhang, Norma Neff, Mark Kowarsky, Christine
  Caneda, et~al.
\newblock Single-cell rna-seq analysis of infiltrating neoplastic cells at the
  migrating front of human glioblastoma.
\newblock \emph{Cell reports}, 21\penalty0 (5):\penalty0 1399--1410, 2017.

\bibitem[Dash and Liu(2000)]{dash2000feature}
Manoranjan Dash and Huan Liu.
\newblock Feature selection for clustering.
\newblock In \emph{Pacific-Asia Conference on knowledge discovery and data
  mining}, pages 110--121. Springer, 2000.

\bibitem[Duarte et~al.(2012)Duarte, Fred, and Duarte]{duarte2012adaptive}
Jo{\~a}o~MM Duarte, Ana~LN Fred, and F~Jorge~F Duarte.
\newblock Adaptive evidence accumulation clustering using the confidence of the
  objects’ assignments.
\newblock In \emph{Pacific-Asia Conference on Knowledge Discovery and Data
  Mining}, pages 70--87. Springer, 2012.

\bibitem[Dudoit and Fridlyand(2003)]{dudoit2003bagging}
Sandrine Dudoit and Jane Fridlyand.
\newblock Bagging to improve the accuracy of a clustering procedure.
\newblock \emph{Bioinformatics}, 19\penalty0 (9):\penalty0 1090--1099, 2003.

\bibitem[Fischer and Buhmann(2003{\natexlab{a}})]{fischer2003bagging}
B.~Fischer and J.M. Buhmann.
\newblock Bagging for path-based clustering.
\newblock \emph{IEEE Transactions on Pattern Analysis and Machine
  Intelligence}, 25\penalty0 (11):\penalty0 1411--1415, 2003{\natexlab{a}}.
\newblock \doi{10.1109/TPAMI.2003.1240115}.

\bibitem[Fischer and Buhmann(2003{\natexlab{b}})]{fischer2003path}
Bernd Fischer and Joachim~M. Buhmann.
\newblock Path-based clustering for grouping of smooth curves and texture
  segmentation.
\newblock \emph{IEEE Transactions on Pattern Analysis and Machine
  Intelligence}, 25\penalty0 (4):\penalty0 513--518, 2003{\natexlab{b}}.

\bibitem[Fred(2001)]{fred2001finding}
Ana Fred.
\newblock Finding consistent clusters in data partitions.
\newblock In \emph{International Workshop on Multiple Classifier Systems},
  pages 309--318. Springer, 2001.

\bibitem[Fred and Jain(2002{\natexlab{a}})]{fred2002evidence}
Ana Fred and Anil~K Jain.
\newblock Evidence accumulation clustering based on the k-means algorithm.
\newblock In \emph{Joint IAPR International Workshops on Statistical Techniques
  in Pattern Recognition (SPR) and Structural and Syntactic Pattern Recognition
  (SSPR)}, pages 442--451. Springer, 2002{\natexlab{a}}.

\bibitem[Fred and Jain(2002{\natexlab{b}})]{fred2002data}
Ana~LN Fred and Anil~K Jain.
\newblock Data clustering using evidence accumulation.
\newblock In \emph{Object recognition supported by user interaction for service
  robots}, volume~4, pages 276--280. IEEE, 2002{\natexlab{b}}.

\bibitem[Fred and Jain(2005)]{fred2005combining}
Ana~LN Fred and Anil~K Jain.
\newblock Combining multiple clusterings using evidence accumulation.
\newblock \emph{IEEE transactions on pattern analysis and machine
  intelligence}, 27\penalty0 (6):\penalty0 835--850, 2005.

\bibitem[Ghaemi et~al.(2009)Ghaemi, Sulaiman, Ibrahim, Mustapha,
  et~al.]{ghaemi2009survey}
Reza Ghaemi, Md~Nasir Sulaiman, Hamidah Ibrahim, Norwati Mustapha, et~al.
\newblock A survey: clustering ensembles techniques.
\newblock \emph{World Academy of Science, Engineering and Technology},
  50:\penalty0 636--645, 2009.

\bibitem[Hayes et~al.(2006)Hayes, Monti, Parmigiani, Gilks, Naoki,
  Bhattacharjee, Socinski, Perou, and Meyerson]{hayes2006gene}
D~Neil Hayes, Stefano Monti, Giovanni Parmigiani, C~Blake Gilks, Katsuhiko
  Naoki, Arindam Bhattacharjee, Mark~A Socinski, Charles Perou, and Matthew
  Meyerson.
\newblock Gene expression profiling reveals reproducible human lung
  adenocarcinoma subtypes in multiple independent patient cohorts.
\newblock \emph{Journal of Clinical Oncology}, 24\penalty0 (31):\penalty0
  5079--5090, 2006.

\bibitem[Karypis and Kumar(1998)]{karypis1998fast}
George Karypis and Vipin Kumar.
\newblock A fast and high quality multilevel scheme for partitioning irregular
  graphs.
\newblock \emph{SIAM Journal on scientific Computing}, 20\penalty0
  (1):\penalty0 359--392, 1998.

\bibitem[Kellam et~al.(2001)Kellam, Liu, Martin, Orengo, Swift, and
  Tucker]{kellam2001comparing}
Paul Kellam, Xiaohui Liu, Nigel Martin, Christine Orengo, Stephen Swift, and
  Allan Tucker.
\newblock Comparing, contrasting and combining clusters in viral gene
  expression data.
\newblock In \emph{Proceedings of 6th workshop on intelligent data analysis in
  medicine and pharmocology}, pages 56--62, 2001.

\bibitem[Kiselev et~al.(2017)Kiselev, Kirschner, Schaub, Andrews, Yiu, Chandra,
  Natarajan, Reik, Barahona, Green, et~al.]{kiselev2017sc3}
Vladimir~Yu Kiselev, Kristina Kirschner, Michael~T Schaub, Tallulah Andrews,
  Andrew Yiu, Tamir Chandra, Kedar~N Natarajan, Wolf Reik, Mauricio Barahona,
  Anthony~R Green, et~al.
\newblock Sc3: consensus clustering of single-cell rna-seq data.
\newblock \emph{Nature methods}, 14\penalty0 (5):\penalty0 483--486, 2017.

\bibitem[Kiselev et~al.(2019)Kiselev, Andrews, and
  Hemberg]{kiselev2019challenges}
Vladimir~Yu Kiselev, Tallulah~S Andrews, and Martin Hemberg.
\newblock Challenges in unsupervised clustering of single-cell rna-seq data.
\newblock \emph{Nature Reviews Genetics}, 20\penalty0 (5):\penalty0 273--282,
  2019.

\bibitem[Liu et~al.(2018)Liu, Shao, and Fu]{liu2018feature}
Hongfu Liu, Ming Shao, and Yun Fu.
\newblock Feature selection with unsupervised consensus guidance.
\newblock \emph{IEEE Transactions on Knowledge and Data Engineering},
  31\penalty0 (12):\penalty0 2319--2331, 2018.

\bibitem[Luo et~al.(2006)Luo, Jing, and Xie]{luo2006combining}
Huilan Luo, Furong Jing, and Xiaobing Xie.
\newblock Combining multiple clusterings using information theory based genetic
  algorithm.
\newblock In \emph{2006 International Conference on Computational Intelligence
  and Security}, volume~1, pages 84--89. IEEE, 2006.

\bibitem[Monti et~al.(2003)Monti, Tamayo, Mesirov, and
  Golub]{monti2003consensus}
Stefano Monti, Pablo Tamayo, Jill Mesirov, and Todd Golub.
\newblock Consensus clustering: a resampling-based method for class discovery
  and visualization of gene expression microarray data.
\newblock \emph{Machine learning}, 52\penalty0 (1-2):\penalty0 91--118, 2003.

\bibitem[Murtagh(1983)]{murtagh1983survey}
Fionn Murtagh.
\newblock A survey of recent advances in hierarchical clustering algorithms.
\newblock \emph{The computer journal}, 26\penalty0 (4):\penalty0 354--359,
  1983.

\bibitem[Nene et~al.(1996)Nene, Nayar, and Murase]{Nene96columbiaobject}
Sameer~A. Nene, Shree~K. Nayar, and Hiroshi Murase.
\newblock Columbia object image library (coil-20).
\newblock Technical report, 1996.

\bibitem[Ng et~al.(2001)Ng, Jordan, and Weiss]{ng2001spectral}
Andrew Ng, Michael Jordan, and Yair Weiss.
\newblock On spectral clustering: Analysis and an algorithm.
\newblock \emph{Advances in neural information processing systems},
  14:\penalty0 849--856, 2001.

\bibitem[Pakhira(2014)]{pakhira2014linear}
Malay~K Pakhira.
\newblock A linear time-complexity k-means algorithm using cluster shifting.
\newblock In \emph{2014 International Conference on Computational Intelligence
  and Communication Networks}, pages 1047--1051. IEEE, 2014.

\bibitem[Parvin et~al.(2013)Parvin, Minaei-Bidgoli, Alinejad-Rokny, and
  Punch]{parvin2013data}
Hamid Parvin, Behrouz Minaei-Bidgoli, Hamid Alinejad-Rokny, and William~F
  Punch.
\newblock Data weighing mechanisms for clustering ensembles.
\newblock \emph{Computers \& Electrical Engineering}, 39\penalty0 (5):\penalty0
  1433--1450, 2013.

\bibitem[Pont{\'e}n et~al.(2008)Pont{\'e}n, Jirstr{\"o}m, and
  Uhlen]{ponten2008human}
Fredrik Pont{\'e}n, Karin Jirstr{\"o}m, and Matthias Uhlen.
\newblock The human protein atlas—a tool for pathology.
\newblock \emph{The Journal of Pathology: A Journal of the Pathological Society
  of Great Britain and Ireland}, 216\penalty0 (4):\penalty0 387--393, 2008.

\bibitem[Ren et~al.(2017)Ren, Domeniconi, Zhang, and Yu]{ren2017weighted}
Yazhou Ren, Carlotta Domeniconi, Guoji Zhang, and Guoxian Yu.
\newblock Weighted-object ensemble clustering: methods and analysis.
\newblock \emph{Knowledge and Information Systems}, 51\penalty0 (2):\penalty0
  661--689, 2017.

\bibitem[Satija et~al.(2015)Satija, Farrell, Gennert, Schier, and
  Regev]{satija2015spatial}
Rahul Satija, Jeffrey~A Farrell, David Gennert, Alexander~F Schier, and Aviv
  Regev.
\newblock Spatial reconstruction of single-cell gene expression data.
\newblock \emph{Nature biotechnology}, 33\penalty0 (5):\penalty0 495--502,
  2015.

\bibitem[Serfling(1974)]{10.1214/aos/1176342611}
R.~J. Serfling.
\newblock {Probability Inequalities for the Sum in Sampling without
  Replacement}.
\newblock \emph{The Annals of Statistics}, 2\penalty0 (1):\penalty0 39 -- 48,
  1974.
\newblock \doi{10.1214/aos/1176342611}.
\newblock URL \url{https://doi.org/10.1214/aos/1176342611}.

\bibitem[Slivkins(2019)]{slivkins2019introduction}
Aleksandrs Slivkins.
\newblock Introduction to multi-armed bandits.
\newblock \emph{arXiv preprint arXiv:1904.07272}, 2019.

\bibitem[Strehl and Ghosh(2002)]{strehl2002cluster}
Alexander Strehl and Joydeep Ghosh.
\newblock Cluster ensembles---a knowledge reuse framework for combining
  multiple partitions.
\newblock \emph{Journal of machine learning research}, 3\penalty0
  (Dec):\penalty0 583--617, 2002.

\bibitem[Toghani and Allen(2021)]{toghani2021mp}
Mohammad~Taha Toghani and Genevera~I Allen.
\newblock Mp-boost: Minipatch boosting via adaptive feature and observation
  sampling.
\newblock In \emph{2021 IEEE International Conference on Big Data and Smart
  Computing (BigComp)}, pages 75--78. IEEE, 2021.

\bibitem[Topchy et~al.(2005)Topchy, Jain, and Punch]{topchy2005clustering}
A.~Topchy, A.K. Jain, and W.~Punch.
\newblock Clustering ensembles: models of consensus and weak partitions.
\newblock \emph{IEEE Transactions on Pattern Analysis and Machine
  Intelligence}, 27\penalty0 (12):\penalty0 1866--1881, 2005.
\newblock \doi{10.1109/TPAMI.2005.237}.

\bibitem[Topchy et~al.(2003)Topchy, Jain, and Punch]{topchy2003combining}
Alexander Topchy, Anil~K Jain, and William Punch.
\newblock Combining multiple weak clusterings.
\newblock In \emph{Third IEEE international conference on data mining}, pages
  331--338. IEEE, 2003.

\bibitem[Topchy et~al.(2004{\natexlab{a}})Topchy, Jain, and
  Punch]{topchy2004mixture}
Alexander Topchy, Anil~K Jain, and William Punch.
\newblock A mixture model for clustering ensembles.
\newblock In \emph{Proceedings of the 2004 SIAM international conference on
  data mining}, pages 379--390. SIAM, 2004{\natexlab{a}}.

\bibitem[Topchy et~al.(2004{\natexlab{b}})Topchy, Minaei-Bidgoli, Jain, and
  Punch]{topchy2004adaptive}
Alexander Topchy, Behrouz Minaei-Bidgoli, Anil~K Jain, and William~F Punch.
\newblock Adaptive clustering ensembles.
\newblock In \emph{Proceedings of the 17th International Conference on Pattern
  Recognition, 2004. ICPR 2004.}, volume~1, pages 272--275. IEEE,
  2004{\natexlab{b}}.

\bibitem[Trapnell et~al.(2014)Trapnell, Cacchiarelli, Grimsby, Pokharel, Li,
  Morse, Lennon, Livak, Mikkelsen, and Rinn]{trapnell2014dynamics}
Cole Trapnell, Davide Cacchiarelli, Jonna Grimsby, Prapti Pokharel, Shuqiang
  Li, Michael Morse, Niall~J Lennon, Kenneth~J Livak, Tarjei~S Mikkelsen, and
  John~L Rinn.
\newblock The dynamics and regulators of cell fate decisions are revealed by
  pseudotemporal ordering of single cells.
\newblock \emph{Nature biotechnology}, 32\penalty0 (4):\penalty0 381--386,
  2014.

\bibitem[Verhaak et~al.(2010)Verhaak, Hoadley, Purdom, Wang, Qi, Wilkerson,
  Miller, Ding, Golub, Mesirov, et~al.]{verhaak2010integrated}
Roel~GW Verhaak, Katherine~A Hoadley, Elizabeth Purdom, Victoria Wang, Yuan Qi,
  Matthew~D Wilkerson, C~Ryan Miller, Li~Ding, Todd Golub, Jill~P Mesirov,
  et~al.
\newblock Integrated genomic analysis identifies clinically relevant subtypes
  of glioblastoma characterized by abnormalities in pdgfra, idh1, egfr, and
  nf1.
\newblock \emph{Cancer cell}, 17\penalty0 (1):\penalty0 98--110, 2010.

\bibitem[Wang et~al.(2018)Wang, Zhang, Sun, and Fang]{wang2018sparse}
Binhuan Wang, Yilong Zhang, Will~Wei Sun, and Yixin Fang.
\newblock Sparse convex clustering.
\newblock \emph{Journal of Computational and Graphical Statistics}, 27\penalty0
  (2):\penalty0 393--403, 2018.

\bibitem[Wang and Allen(2021)]{wang2021integrative}
Minjie Wang and Genevera~I Allen.
\newblock Integrative generalized convex clustering optimization and feature
  selection for mixed multi-view data.
\newblock \emph{Journal of Machine Learning Research}, 22\penalty0
  (55):\penalty0 1--73, 2021.

\bibitem[Watanabe et~al.(2002)Watanabe, Maemura, Kanbara, Tamayama, and
  Hayasaki]{watanabe2002gaba}
Masahito Watanabe, Kentaro Maemura, Kiyoto Kanbara, Takumi Tamayama, and Hana
  Hayasaki.
\newblock Gaba and gaba receptors in the central nervous system and other
  organs.
\newblock \emph{International review of cytology}, 213:\penalty0 1--47, 2002.

\bibitem[Weinstein et~al.(2013)Weinstein, Collisson, Mills, Shaw, Ozenberger,
  Ellrott, Shmulevich, Sander, and Stuart]{weinstein2013cancer}
John~N Weinstein, Eric~A Collisson, Gordon~B Mills, Kenna R~Mills Shaw, Brad~A
  Ozenberger, Kyle Ellrott, Ilya Shmulevich, Chris Sander, and Joshua~M Stuart.
\newblock The cancer genome atlas pan-cancer analysis project.
\newblock \emph{Nature genetics}, 45\penalty0 (10):\penalty0 1113--1120, 2013.

\bibitem[Wilkerson and Hayes(2010)]{wilkerson2010consensusclusterplus}
Matthew~D Wilkerson and D~Neil Hayes.
\newblock Consensusclusterplus: a class discovery tool with confidence
  assessments and item tracking.
\newblock \emph{Bioinformatics}, 26\penalty0 (12):\penalty0 1572--1573, 2010.

\bibitem[Witten and Tibshirani(2010)]{witten2010framework}
Daniela~M Witten and Robert Tibshirani.
\newblock A framework for feature selection in clustering.
\newblock \emph{Journal of the American Statistical Association}, 105\penalty0
  (490):\penalty0 713--726, 2010.

\bibitem[Wolf et~al.(2018)Wolf, Angerer, and Theis]{wolf2018scanpy}
F~Alexander Wolf, Philipp Angerer, and Fabian~J Theis.
\newblock Scanpy: large-scale single-cell gene expression data analysis.
\newblock \emph{Genome biology}, 19\penalty0 (1):\penalty0 1--5, 2018.

\bibitem[Yang et~al.(2019)Yang, Huh, Culpepper, Lin, Love, and
  Li]{yang2019safe}
Yuchen Yang, Ruth Huh, Houston~W Culpepper, Yuan Lin, Michael~I Love, and Yun
  Li.
\newblock Safe-clustering: single-cell aggregated (from ensemble) clustering
  for single-cell rna-seq data.
\newblock \emph{Bioinformatics}, 35\penalty0 (8):\penalty0 1269--1277, 2019.

\bibitem[Yao and Allen(2020)]{yao2020feature}
Tianyi Yao and Genevera~I Allen.
\newblock Feature selection for huge data via minipatch learning.
\newblock \emph{arXiv preprint arXiv:2010.08529}, 2020.

\bibitem[Yao et~al.(2021)Yao, LeJeune, Javadi, Baraniuk, and
  Allen]{yao2021minipatch}
Tianyi Yao, Daniel LeJeune, Hamid Javadi, Richard~G Baraniuk, and Genevera~I
  Allen.
\newblock Minipatch learning as implicit ridge-like regularization.
\newblock In \emph{2021 IEEE International Conference on Big Data and Smart
  Computing (BigComp)}, pages 65--68. IEEE, 2021.

\bibitem[Yu et~al.(2019)Yu, Zhong, and Kim]{yu2019ensemble}
Jaehong Yu, Hua Zhong, and Seoung~Bum Kim.
\newblock An ensemble feature ranking algorithm for clustering analysis.
\newblock \emph{Journal of Classification}, pages 1--28, 2019.

\bibitem[Zhao and Liu(2007)]{zhao2007spectral}
Zheng Zhao and Huan Liu.
\newblock Spectral feature selection for supervised and unsupervised learning.
\newblock In \emph{Proceedings of the 24th international conference on Machine
  learning}, pages 1151--1157, 2007.

\end{thebibliography}

\end{document}


\maketitle

\appendix

\section{Proof of Proporsition~\ref{le:tc}}\label{sec:pp}
\begin{prop}\label{le:tc}  
The computational complexity of MPCC in Algorithm~1 is $O(mn^2T+N^2)$, where $T$ is the total number of minipatches.
\end{prop}
\begin{proof}
Firstly, construction of minipatch requires $O(1)$ time complexity. Then we need $O(\frac{mn(n-1)}{2})$ to calculate the distance matrix of minipatch and $O(n^2)$ to perform hierarchical clustering with Ward.D linkage [Murtagh,1983]. The time complexity is $O(\frac{n(n-1)}{2})$ for both co-clustering membership matrix $\boldsymbol{\mathcal{V}}$ and co-sampling matrix $\boldsymbol{\mathcal{D}}$ updates. Finally, after $T$ iterations, when we calculate the consensus matrix $\boldsymbol{\mathcal{S}}$, $O(N^2)$ operations are needed. Therefore, the overall time complexity over $T$ iterations is $O((mn(n-1)+n^2+n(n-1))\times T+N^2)$, which can be simplified to $O(mn^2T+N^2)$. 
\end{proof}

\section{Early Stopping Criteria}\label{sec:early}

To avoid unnecessary iterations in an effort to optimize computational efficiency, we employ an early stopping criteria which stops the algorithm once the consensus matrix is stable. The stability can be measured by the changes of confusion values [Ren et al., 2017], where the confusion is defined as  
$$
    confusion_i = \frac{1}{N}\sum_{i'=1}^N\boldsymbol{\mathcal{S}}_{i,i'}(1-\boldsymbol{\mathcal{S}})_{i,i'}
$$
for observation $i$. We consider the consensus results to be stable if the confusion values have no or little changes over several iterations.  Let $\epsilon_{q^{th}}^{(t)}$ be the absolute difference between the $q^{th}$ percentile of confusion values at iteration $t$ and $t-1$; and $\Omega_{q,c}$ denotes the number of times that $\epsilon_{q^{th}}$ is continuously less than a constant $c$. We propose to stop the algorithm when the $90\%$ percentile value of the confusions remains small and unchanged for the past $5$ iterations, that it, $\Omega_{90,0.00001}<5$ for $c = 0.00001$. Such stopping criteria is used in our empirical study and proved to achieve a great balance between clustering accuracy and computational efficiency.

\section{Additional Study on Synthetic Data}\label{sec:moresimu}
\subsection{Weak Sparse Simulation}

    In order to further investigate the effectiveness of MPCC and IMPACC methods, we propose to evaluate the methods on more difficult and more noisy data sets. To do so, we construct a weak sparse scenario, in which we generate noise features with higher variance than that of sparse simulation as in Section~3.1. Instead of setting the means of all noise features to be $0$, here, we design the weak sparse simulation with means of noise features $\mu_0\sim N(\textbf{0}_{4975},I_{4975})$, where $I$ is the identity matrix.  With other settings being exactly the same as sparse simulation in Section~3.1, we compare model performance of IMPACC and MPCC to other competing methods. As shown in Figure~\ref{fig:simu_weak}, the results of weak sparse scenario are quite similar to that in sparse simulation, with IMPACC outperforms all other methods in terms of clustering accuracy, measured by ARI. Also, IMPACC demonstrates better feature selection accuracy measured by F1 score and much fewer computation costs than other sparse clustering methods.

\setcounter{figure}{2}    

\begin{figure}[!h]
\centering\includegraphics[scale=0.06]{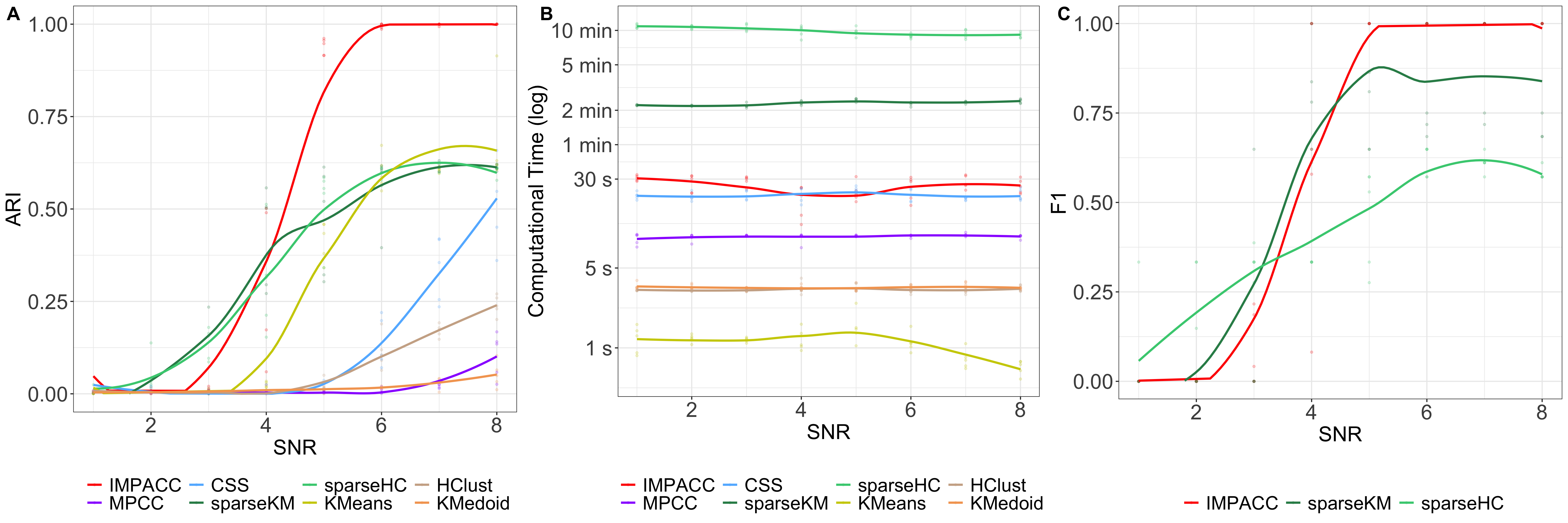}
\caption{Clustering performance (ARI), feature selection accuracy (F1 score), and computation time on sparse synthetic data sets. (A) ARI (higher is better) of estimated grouping; (B) computation time in log seconds; (C) F1 score for signal feature estimates. The performance is quite similar to sparse simulation that our method IMPACC has superior performance over competing methods in terms of clustering performance, and similar feature selection accuracy with significant improvement on computational cost. }\label{fig:simu_weak}
\end{figure}

\subsection{No Sparse Simulation}

\begin{figure}[!h]
\centering\includegraphics[scale=0.08]{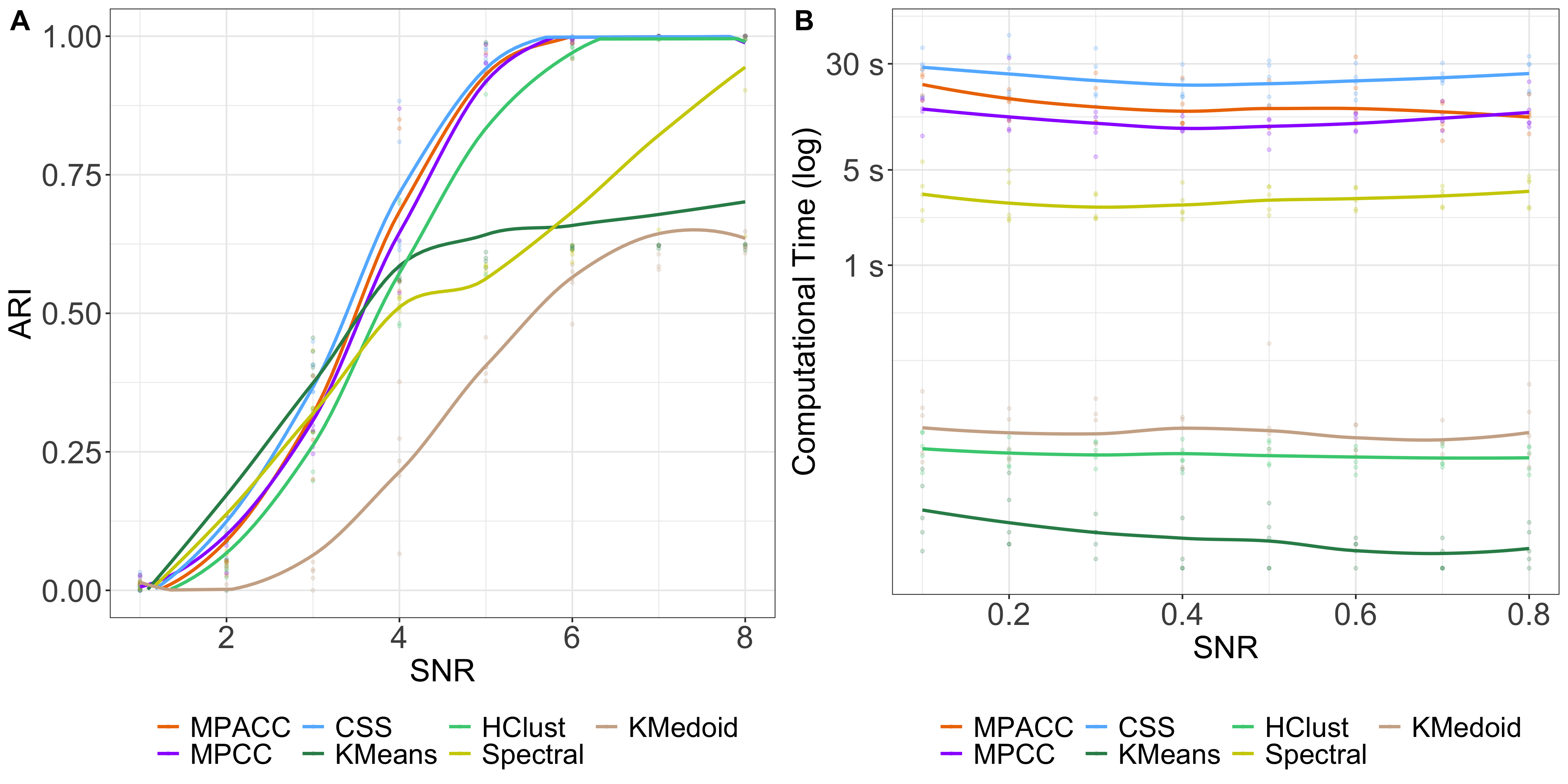}
\caption{Clustering performance (ARI) and computation time on no sparsity and low dimensional synthetic data sets.(A) Adjusted rand index (higher is better) of estimated grouping; (B) computation time in log seconds; Our method MPACC and MPCC have similar clustering accuracy and advantages in computation time comparing to consensus clustering. They also demonstrate much higher ARI over other competing methods.}\label{fig:nosparse}
\end{figure}

As our previous synthetic data sets are built with sparse and high dimensional settings, one may ask how do our methods perform in the case of no sparsity and low dimensions?  For comparison, we generate a low dimensional data sets that the number of features ($M = 100$) is smaller than the number of observations($N = 500$), with all the features being relevant to differentiate the clusters. Similar to the sparse scenarios, the no sparse data sets are simulated from a mixture of Gaussian with block-diagonal covariance matrix. Differently, here, the number of features is set to be $M = 100$, and other values are the same as in sparse simulation. Without any noise feature, the $100$ signal features are generated with $\mu = [\mu_{k}]$, such that $\mu_k\sim N(u_k,0.1)$, and we set $u_1 = \frac{SNR}{10}\cdot\textbf{1}_{100}$, $\mu_2 = (\frac{SNR}{10}\cdot\textbf{1}_{50}^T, -\frac{SNR}{10}\cdot\textbf{1}_{50}^T)^T$, $\mu_3 = (-\frac{SNR}{10}\cdot\textbf{1}_{50}^T, \frac{SNR}{10}\cdot\textbf{1}_{50}^T)^T$, $\mu_4 = -\frac{SNR}{10}\cdot\textbf{1}_{50}$. In this case, all features are important in discovering clusters, so we do not implement clustering algorithms with feature selection, including IMPACC, sparseKM and sparseHC. We aim to compare MPACC, which only adaptively subsamples observations, MPCC, which uses a random sampling scheme, to consensus clustering and other widely used algorithms. As shown in Figure~\ref{fig:nosparse}, we can tell that MPACC and MPCC do not offer significant improvements in terms of clustering accuracy, comparing to consensus clustering in the no sparsity setting. But our methods show better computational efficiency than consensus clustering. Besides, MPACC and MPCC still result in much better clustering accuracy than other standard clustering algorithms including K-Means, K-Medoid and spectral clustering.

\subsection{Additional Results on Sparse Simulation}

\begin{figure}[!h]
\centering\includegraphics[scale=0.1]{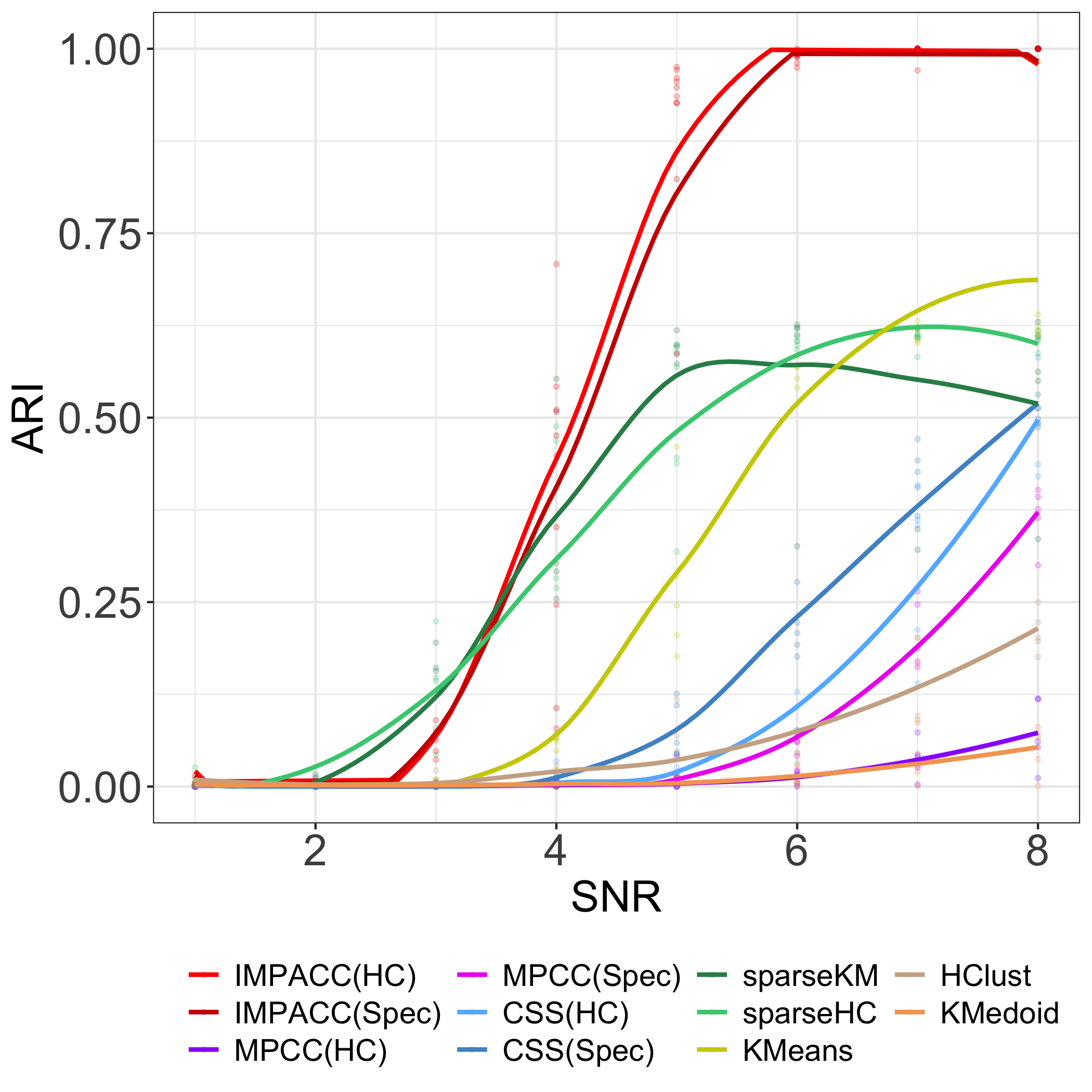}
\caption{Clustering performance (ARI) on sparse data sets, adding results of spectral clustering as the final algorithm. Methods using hierarchical and spectral clustering are indicated by (HC) and (Spec), respectively. IMPACC has similar ARI in both \textit{HC} and \textit{Spec} settings, while MPCC and consensus clustering with \textit{Spec} show better clustering accuracy than that with \textit{HC} setting.}\label{fig:simu_al}
\end{figure}

One may ask can we implement algorithms other than hierarchical clustering on the consensus matrix in the final step, and whether the methods would perform differently? In practice, our frameworks are quite flexible and we can implement any other clustering algorithms in the final step, as long as they rely on a distance matrix or similarity matrix of the data. For example, we can apply hierarchical clustering or K-Medoid clustering with $1-\boldsymbol{\mathcal{S}}$ as a distance matrix, or spectral clustering using $\boldsymbol{\mathcal{S}}$ as a similarity matrix. In the setting of sparse simulation scenario, we compare clustering accuracy of our frameworks to existing methods, adding spectral clustering as the final algorithm in IMPACC, MPCC and consensus clustering. As shown in Figure~\ref{fig:simu_al}, the methods using hierarchical and spectral clustering as the final algorithm are indicated by \textit{(HC)} and \textit{(Spec)}, respectively. Since the only difference is the final clustering applied to the consensus matrix, only the clustering accuracy would be influenced. IMPACC with spectral clustering as final algorithm \textit{IMAPCC(Spec)} has almost the same ARI as that of using hierarchical clustering as final algorithm \textit{IMAPCC(HC)}. And in both MPCC and consensus clustering, spectral clustering as the final algorithm significantly enhances clustering accuracy. Therefore, our approaches are flexible and widely applicable to numerous of different models in the choices of final clustering algorithms, so that users can optimize model settings with respect to data of interest.

\subsubsection{Patterns of Feature Scores}

\begin{figure}[!t]
\centering\includegraphics[scale=0.15]{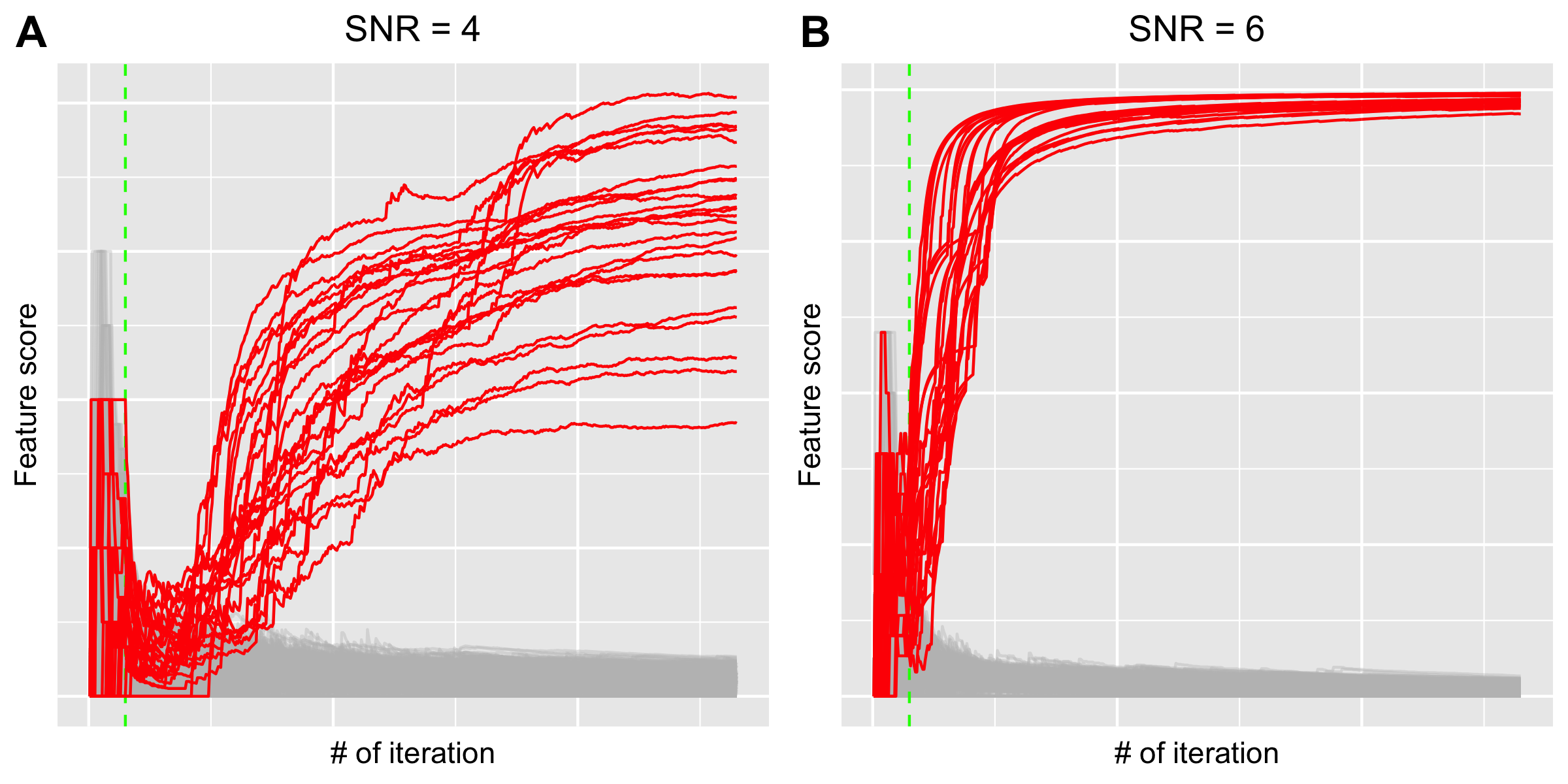}
\caption{Feature scores of IMAPCC versus number of iterations in the sparse scenario with $SNR = 4$ and $SNR = 6$. Our method is able to distinguish signal features from the noise, with speed of convergence and level of score separation relative to the difficulty of the data sets. }\label{fig:impo}
\end{figure}

One may also be interested in the pattern of feature scores over iterations. We display feature scores obtained from IMPACC versus the number of iterations in the sparse synthetic data study, in the case of $SNR = 4$ and $SNR = 6$. From Figure~\ref{fig:impo}, we see that IMPACC can correctly identify all the signal features and noise features after the burn-in stage. In addition, the feature scores converge faster and have a clearer separation between the important and noise features in the data with higher $SNR$. It is reasonable because the important features in such data have stronger signals, so it is easier for us to separate the relevant features from the noisy ones, and thus easier to cluster the observations.  

\section{Important Genes in Genomics Data Sets}\label{sec:gene}

Our method IMPACC can in addition generate interpretable results with ensembled feature selection. In the RNA-seq data sets of our real data case study, we are able to find biologically important genes by the ranks of resulting feature importance. With gene information extracted from the Human Protein Atlas~[Pontén et al., 2008], we list the top $25$ genes with the highest importance score, and the descriptions of their tissue specificity and/or whether they are prognostic cancer markers in Table~\ref{table:gene_pancan},~\ref{table:gene_brain},~\ref{table:gene_neoplastic}. In addition, Table~\ref{table:path_p} and Table~\ref{table:path_n} contain details on top 5 enriched pathways of signal genes detected by IMPACC and sparseKM in the PANCAN tumor data set and neoplastic data set. From the information, we can tell that IMAPCC can effectively identify influential genes and enriched pathways that are actually biologically meaningful, supported by the scientific literature.

\begin{table}[h]
\centering
\resizebox{\textwidth}{!}{\begin{tabular}{lrrrrrrr}
\toprule
\multicolumn{1}{c}{ } & \multicolumn{3}{c}{\bf IMAPCC} & \multicolumn{3}{c}{\bf sparseKM } \\
\cmidrule(l{3pt}r{3pt}){2-4} \cmidrule(l{3pt}r{3pt}){5-7} 
  & Pathway & Name  & p-value & Pathway & Name  & p-value\\
\midrule
1 & hsa05217 & Basal cell carcinoma & 0.00225 & hsa00250 & Alanine, aspartate and glutamate metabolism & 0.0122\\
2 & hsa05224 & Breast cancer & 0.0058& hsa04080 & Neuroactive ligand-receptor interaction & 0.0185\\
3 & hsa00601 & Glycosphingolipid biosynthesis - lacto and neolacto series & 0.0070& hsa04520 & Adherens junction & 0.0415\\
4 & hsa05165 & Human papillomavirus infection & 0.0093 & hsa04918 & Thyroid hormone synthesis & 0.0458\\
5& hsa05166 & Human T-cell leukemia virus 1 infection & 0.0164 & hsa05016 & Huntington disease & 0.0497\\
\bottomrule

\end{tabular}}
\caption{Top 5 enriched KEGG pathways in PANCAN tumor data set}
\label{table:path_p}
\end{table}

\begin{table}[h]
\centering
\resizebox{\textwidth}{!}{\begin{tabular}{lrrrrrrr}
\toprule
\multicolumn{1}{c}{ } & \multicolumn{3}{c}{\bf IMAPCC} & \multicolumn{3}{c}{\bf sparseKM } \\
\cmidrule(l{3pt}r{3pt}){2-4} \cmidrule(l{3pt}r{3pt}){5-7} 
  & Pathway & Name  & p-value & Pathway & Name  & p-value\\
\midrule

1 & hsa04145 & Phagosome & 7.193e-14 & hsa05150 & Staphylococcus aureus infection & 4.551e-17\\
2& hsa05150 & Staphylococcus aureus infection & 1.006e-13 & hsa04145 & Phagosome & 2.351e-15\\
3 & hsa04514 & Cell adhesion molecules & 2.0269e-13 & hsa05140 & Leishmaniasis & 2.479e-15\\
4& hsa04940 & Type I diabetes mellitus & 8.144e-13 & hsa05152 & Tuberculosis & 6.569e-13\\
5& hsa05332 & Graft-versus-host disease & 6.259e-12 & hsa05322 & Systemic lupus erythematosus & 4.861e-12\\
\bottomrule

\end{tabular}}
\caption{Top 5 enriched KEGG pathways in neoplastic cell data set}
\label{table:path_n}
\end{table}

\begin{table}[h]
\centering
 \begin{tabular}{l p{2cm} p{8cm}}

\toprule
  & Gene name & Description\\
\midrule

  &ACPL2&Tissue enhanced (epididymis)\\
  &   ANXA3&Prognostic marker in renal cancer and pancreatic cancer \\
 & CTXN3 & group enriched (brain) \\
  &   EPB41L3& Tissue enhanced (brain) \\

 &    GATA2&Prognostic marker in urothelial cancer and renal cancer  \\
 
  &   MRE11A&Prognostic marker in liver cancer  \\
  &   MYBPH&Tissue enriched (skeletal muscle) \\
  &   NKX2-8&Tissue enhanced (lung) \\
  &   SLC45A1&Tissue enhanced (brain) \\

  &   TRPM5&Tissue enhanced in pancreas\\
  &   CNIH4&Prognostic marker in renal cancer and pancreatic cancer  \\
 &    MLNR &Group enriched (stomach) \\
  &   SEMA4G&Tissue enhanced (intestine, liver) \\
  &   CREB3L2&Tissue enhanced (placenta)\\
  &   RDBP&Prognostic marker in liver cancer and cervical cancer  \\
  &  ACAD10& Prognostic marker in renal cancer \\
  &   CDH5&Prognostic marker in renal cancer  \\
  &   PVRIG&Group enriched (blood, lymphoid tissue) \\

  &   KCNE2&Tissue enriched (stomach) \\

  &   EFHC2&Tissue enhanced (brain, fallopian tube) \\

  &   ENOX2&Prognostic marker in renal cancer  \\
   &  GALNT4&not prognostic \\
   &  TBC1D7&Prognostic marker in renal cancer  and ovarian cancer  \\
   &  TMEM30C&Prognostic marker in renal cancer  \\
   &   SYT5& Prognostic marker in pancreatic cancer and glioma \\

\bottomrule
\end{tabular}
\caption{Top 25 significant genes derived from IMPACC in PANCAN}
\label{table:gene_pancan}
\end{table}

\begin{table}[h]
\centering
 \begin{tabular}{l p{2cm} p{8cm}}
\toprule
  & Gene name & Description\\
\midrule
  & THY1& Tissue enhanced (brain, smooth muscle) \\
 & TMEM130& Group enriched (brain) \\
 & GABRB2& Tissue enriched (brain) \\
 & AGXT2L1& Group enriched (brain, liver) \\
 & UGT8& Tissue enriched (brain) \\
 & RNASE1& Tissue enriched (pancreas)\\
 & GABRA1& Tissue enriched (brain) \\
 & ERMN& Tissue enriched (brain)\\
 & OPALIN& Tissue enriched (brain) \\
 & KLK6& Tissue enhanced (brain, esophagus) \\
 & CLDN11& Tissue enhanced (brain, ovary, testis) \\
 & SCN2A& Tissue enriched (brain) \\
 & GABBR2& Tissue enriched (brain) \\
 & ENPP2& Tissue enhanced (brain, placenta)\\
 & CNDP1& Group enriched (brain, liver)\\
 & SLCO1A2& Tissue enhanced (brain, liver, retina, salivary gland)\\
 & GJA1& Low tissue specificity\\
 & AQP4& Group enriched (brain, lung)\\
 & BCAS1& Tissue enhanced (brain, intestine, stomach)\\
 & TF& Tissue enriched (liver)\\
 & CLU& Tissue enhanced (epididymis, liver)\\
 & UNC80& Group enriched ( brain)\\
 & SPP1& Tissue enhanced (gallbladder, kidney, placenta)\\
 & PLP1& Tissue enriched (brain)\\

\bottomrule
\end{tabular}
\caption{Top 25 significant genes derived from IMPACC in brain cells}
\label{table:gene_brain}
\end{table}

\begin{table}[h]
\centering
 \begin{tabular}{l p{2cm} p{9cm}}
\toprule
  & Gene name & Description\\
\midrule
& TYROBP& Prognostic marker in renal cancer \\
& CD74& Prognostic marker in endometrial cancer  and breast cancer \\

& PTPRZ1& Tissue enriched (brain); Prognostic marker in urothelial cancer \\

& SRGN& Prognostic marker in renal cancer \\
& FCER1G& Prognostic marker in renal cancer \\
& LAPTM5& Prognostic marker in renal cancer and cervical cancer\\ 
& C1QB& Prognostic marker in renal cancer \\
& HLA-DRB1& Gene product is not prognostic\\
& HLA-DRA& Prognostic marker in colorectal cancer \\
& C1QA& Prognostic marker in renal cancer \\
& C1QC& Prognostic marker in renal cancer \\
& APOD& Prognostic marker in breast cancer  and stomach cancer \\
& HLA-DRB5& Prognostic marker in endometrial cancer \\
& GAD2& Tissue enriched (brain)\\
& SOX10& Tissue enhanced (brain)\\
& GJB6& Gene product is not prognostic\\
& GPM6A& Tissue enriched (brain)\\
& GPM6B& Tissue enriched (brain); Prognostic marker in ovarian cancer \\
& AIF1& Prognostic marker in renal cancer \\
& PTN& Prognostic marker in renal cancer \\
& CALY& Group enriched (adrenal gland); Prognostic marker in pancreatic cancer \\
& BCAN& Tissue enriched (brain)\\ 

& CLU& Prognostic marker in thyroid cancer \\

& SLC32A1& Tissue enriched (brain)\\

& OPALIN& Tissue enriched (brain)\\

\bottomrule
\end{tabular}
\caption{Top 25 significant genes derived from IMPACC in neoplastic cells}
\label{table:gene_neoplastic}
\end{table}

\newpage

\section{Dimension Reduction \& visualization}\label{sec:dr}
\begin{figure}[!t]
\centering
\includegraphics[scale=0.09]{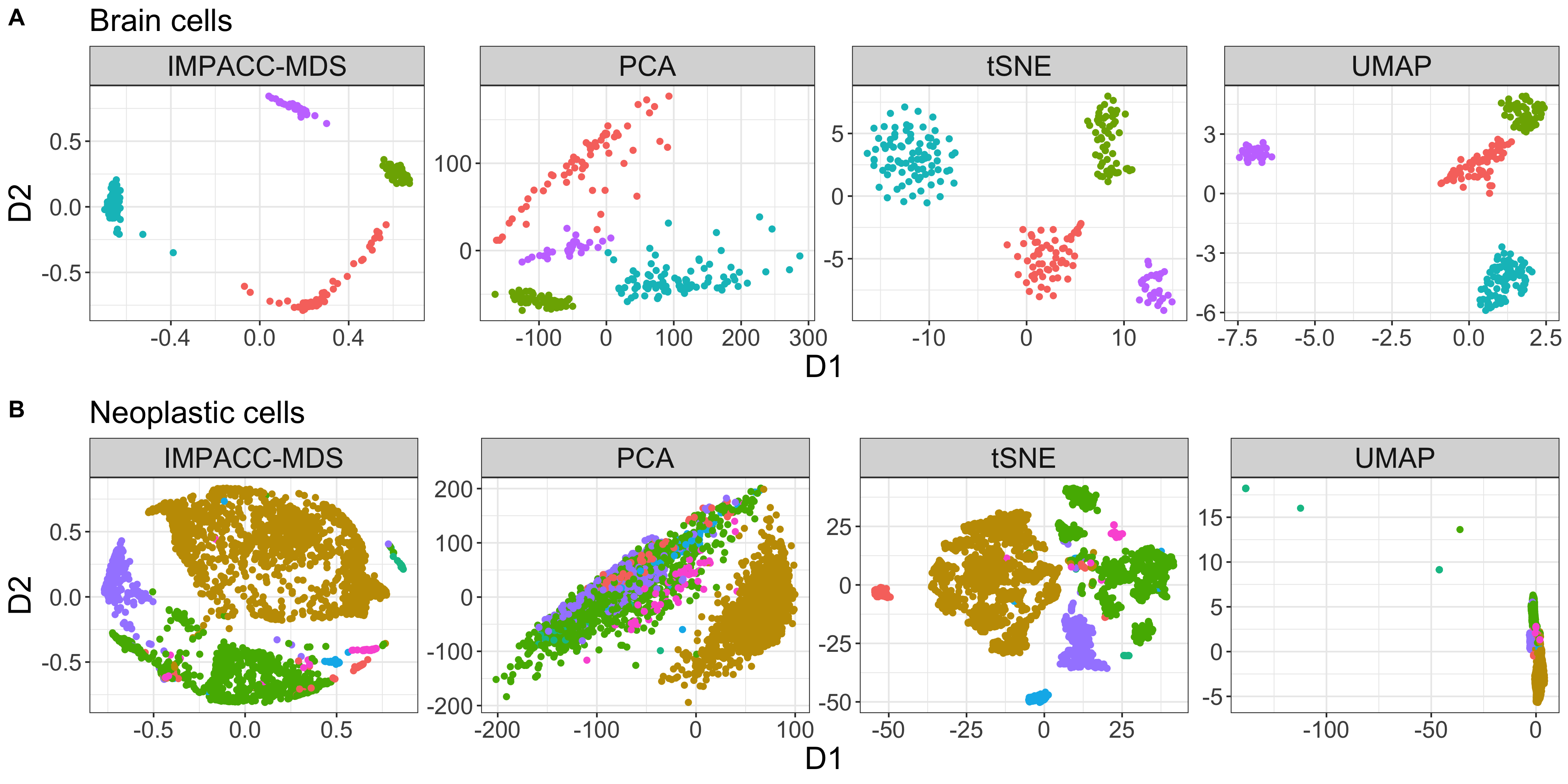}
\caption{Dimension reduction accuracy. Scatterplots of the top two dimension of the cells colored by cell type, derived from MDS-IMPACC, PCA, t-SNE and UMAP. Scatterplots from IMPACC-MDS yield more concentrated clusters with clearer boundaries.}\label{fig:mds}
\end{figure}

Informative dimension reduction and visualization can be achieved by applying multidimensional scaling to the distance matrix obtained from IMPACC, denoted as MDS-IMPACC. The embedding is produced by implementing monotone spline transformed MDS to $1 - \boldsymbol{\mathcal{S}}$, which can be regarded as an interpretable distance matrix estimation of the data. Both the number of knots and spline degree are set to be 2 by default. MDS-IMPACC has significantly better quality of visualization comparing to commonly used dimension reduction techniques including PCA, UMAP and t-SNE. The visualization produced by MDS-IMPACC is equipped with strong interpretability by preserving pairwise distance structure using multidimensional scaling. Moreover, MDS-IMPACC is possible to provide additional information on the levels of clustering uncertainty of the observations, using confusions derived from the final consensus matrix of IMPACC. Figure~\ref{fig:mds} shows the results of applying MDS-IMPACC, PCA, t-SNE and UMAP in the brain cells data and neoplastic cell data set. 
MDS-IMPACC yields scatterplots with tighter groups of cells and clearer separations among different types of cells than PCA and t-SNE. UMAP also constructs competitive visualization in brains cell data, but is sensitive and performs much worse to neoplastic data which has highly unbalanced clusters. Therefore, based on monotone spline multidimensional scaling, IMPACC is able to provide a new interpretable and informative dimension reduction and visualization.

\newpage
\section{A study on Hyper-parameters and Hyper-parameter Tuning}\label{sec:hyper}

\begin{figure}[!h]
\centering\includegraphics[scale=0.08]{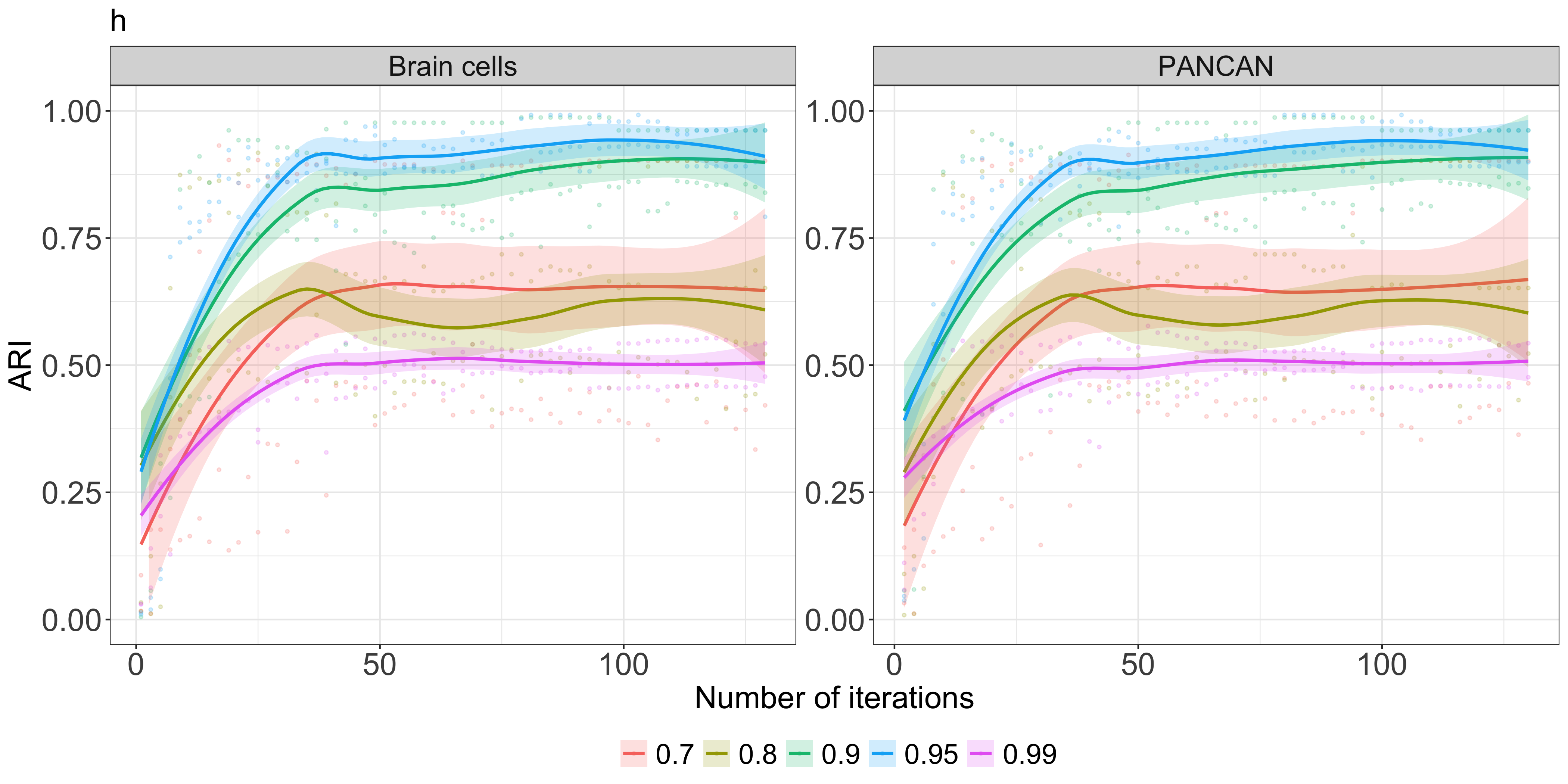}
\caption{Clustering performance measured by ARI with different values of hierarchical tree cutoff $h$ in IMPACC using brain cells and PANCAN data. Methods with $h = 0.95$ setting have significantly better clustering performance. So we use $h=0.95$ as default value.}\label{fig:hyper1}
\end{figure}

\begin{figure}[!h]
\centering\includegraphics[scale=0.08]{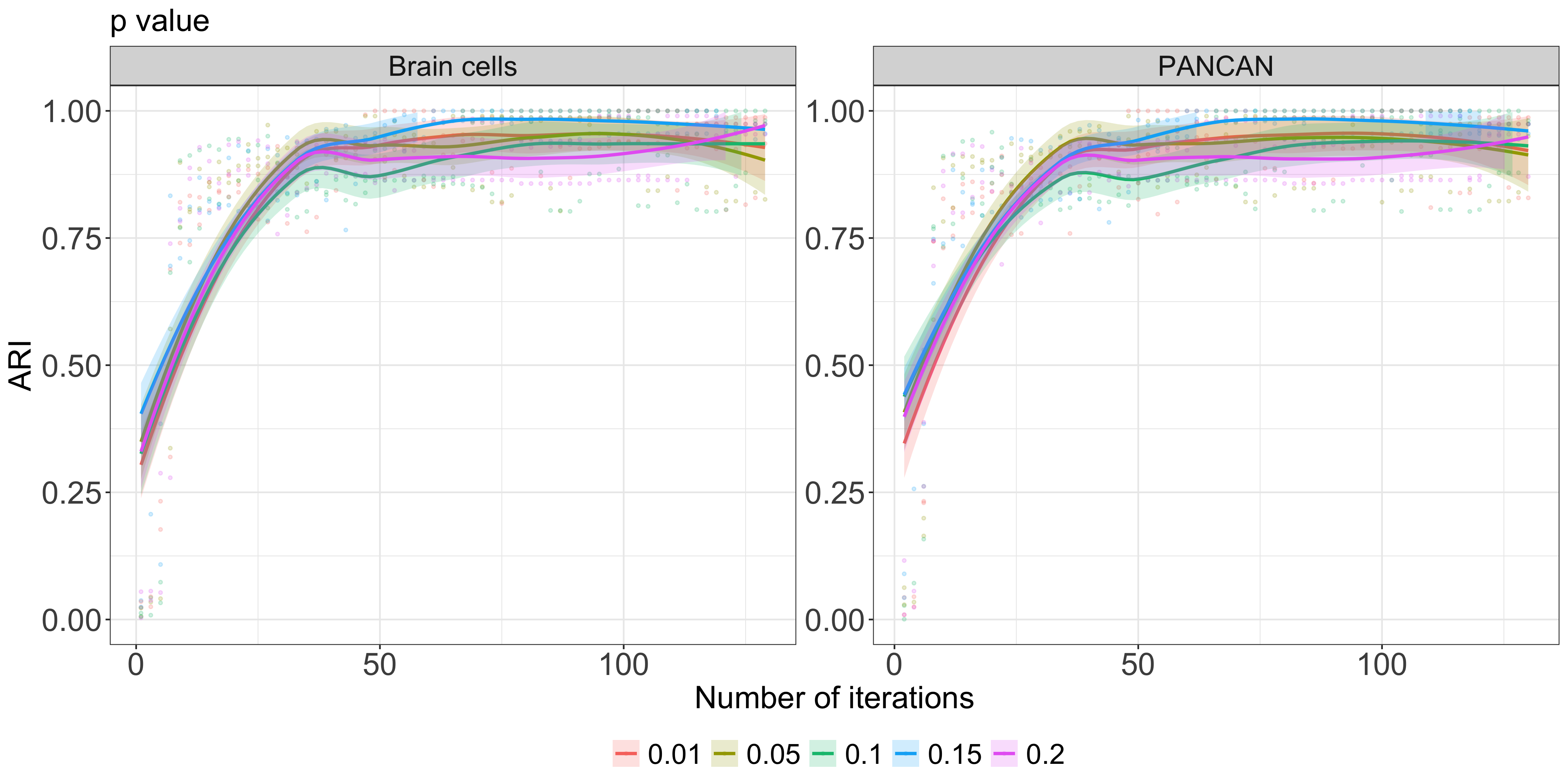}
\caption{Clustering performance measured by ARI with different values of p-value percentile cutoff $\{\eta\}$ in IMPACC using brain cells and PANCAN data. Clustering accuracy is not significantly different with various $\{\eta\}$  settings.}\label{fig:hyper2}
\end{figure}

\begin{figure}[!h]
\centering\includegraphics[scale=0.08]{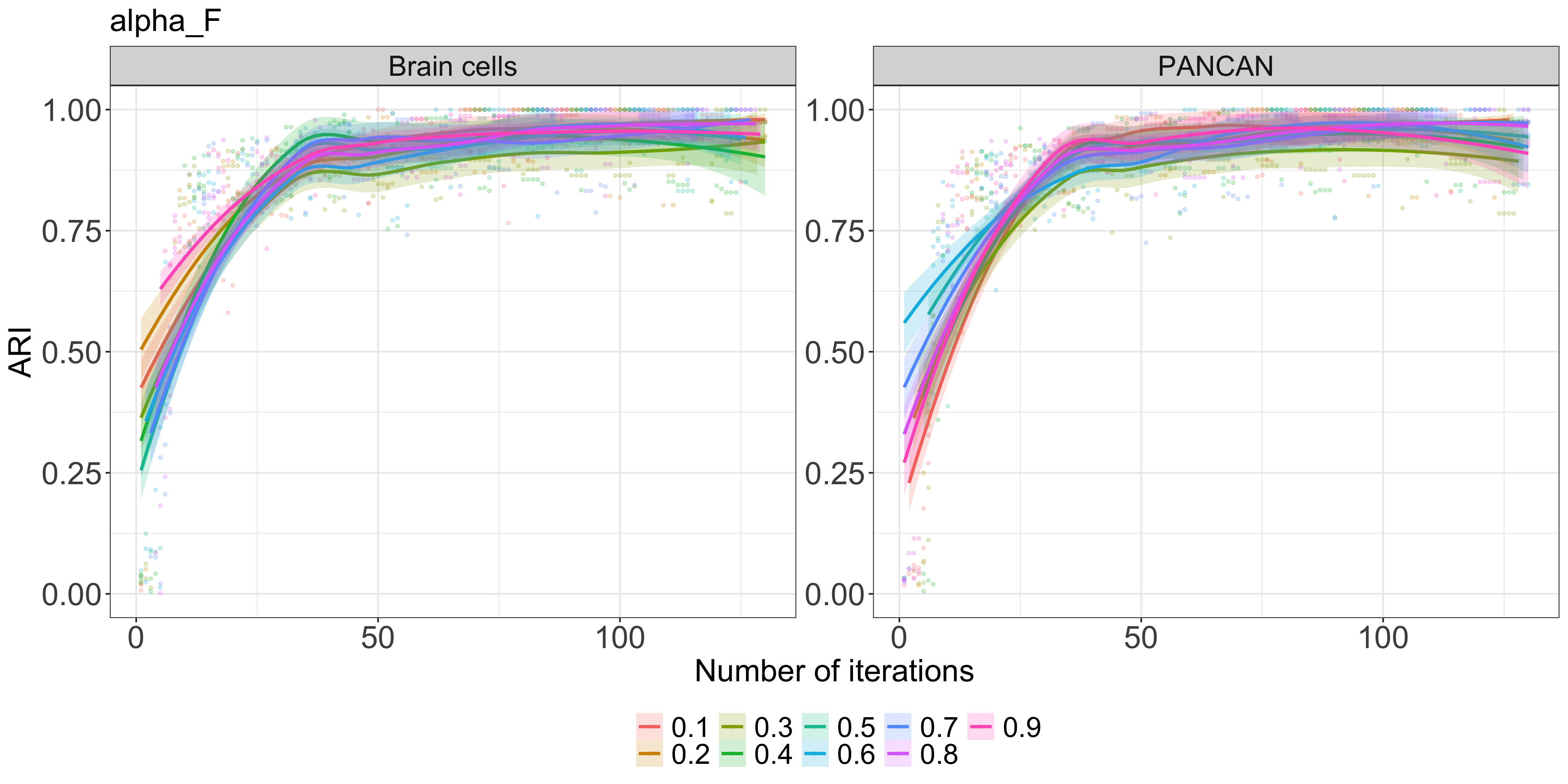}
\caption{Clustering performance measured by ARI with different values of feature learning rate $\alpha_F$ in IMPACC using brain cells and PANCAN data. Clustering accuracy is not significantly different with various $\alpha_F$ settings. }\label{fig:hyper3}
\end{figure}

\begin{figure}[!h]
\centering\includegraphics[scale=0.08]{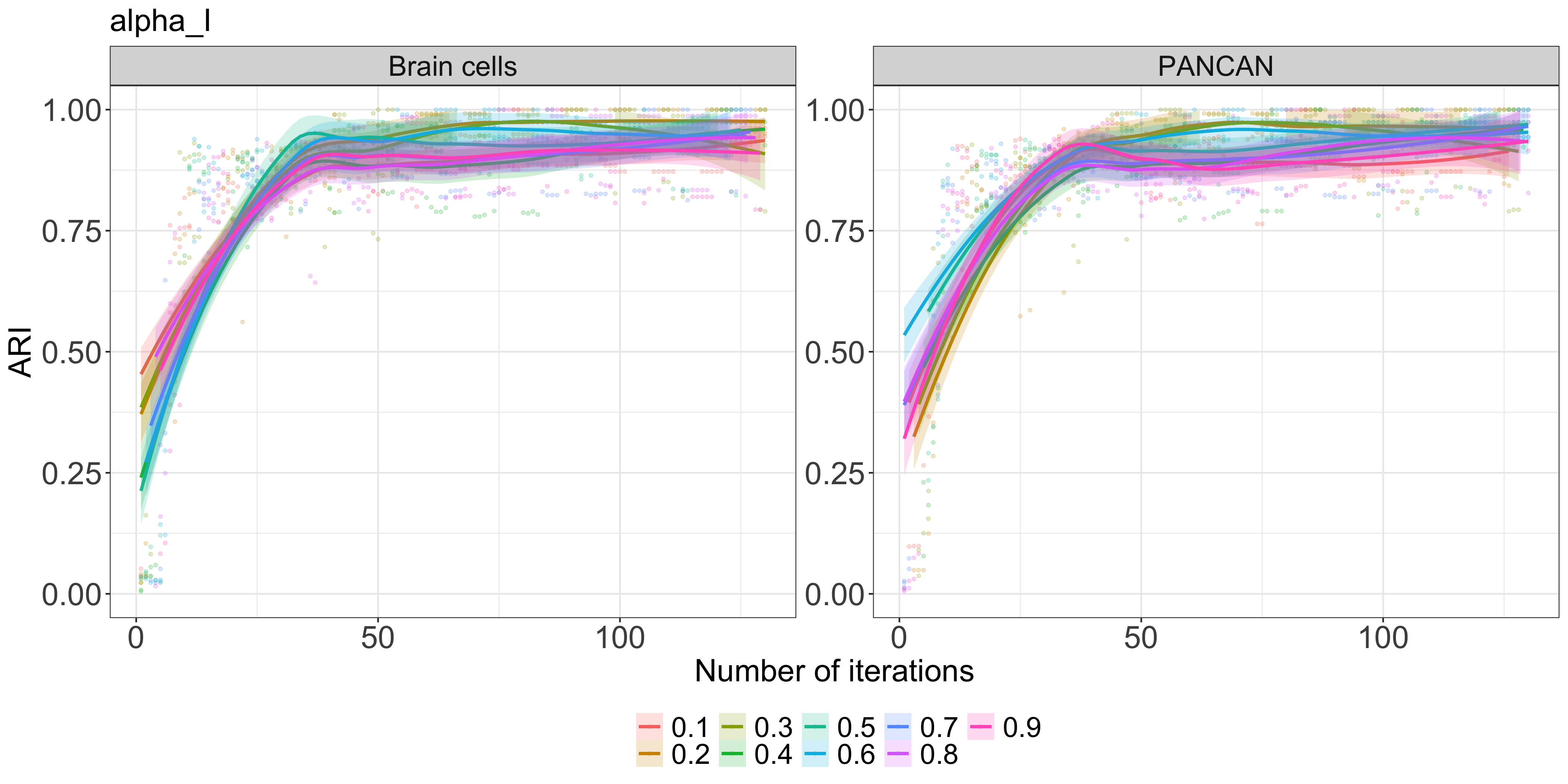}
\caption{Clustering performance measured by ARI with different values of observation learning rate $\alpha_I$ in IMPACC using brain cells and PANCAN data. Clustering accuracy is not significantly different with various $\alpha_I$ settings.}\label{fig:hyper4}
\end{figure}

\begin{figure}[h]
\centering\includegraphics[scale=0.08]{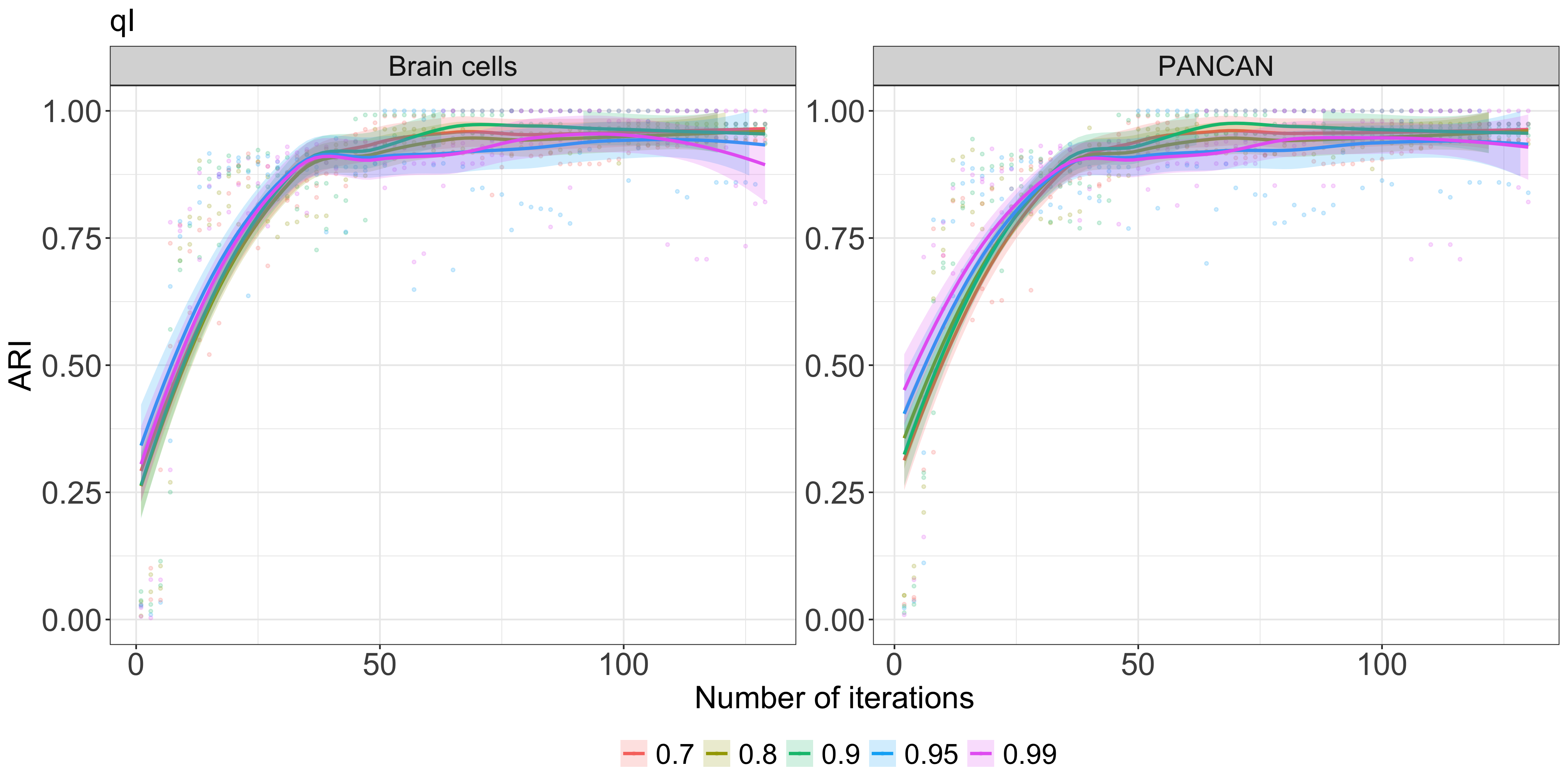}
\caption{Clustering performance measured by ARI with different values of high uncertainty cutoff $\{\theta\}$ in IMPACC using brain cells and PANCAN data. Clustering accuracy is not significantly different with various $\{\theta\}$ settings.}\label{fig:hyper5}
\end{figure}

\begin{figure}[!h]
\centering\includegraphics[scale=0.08]{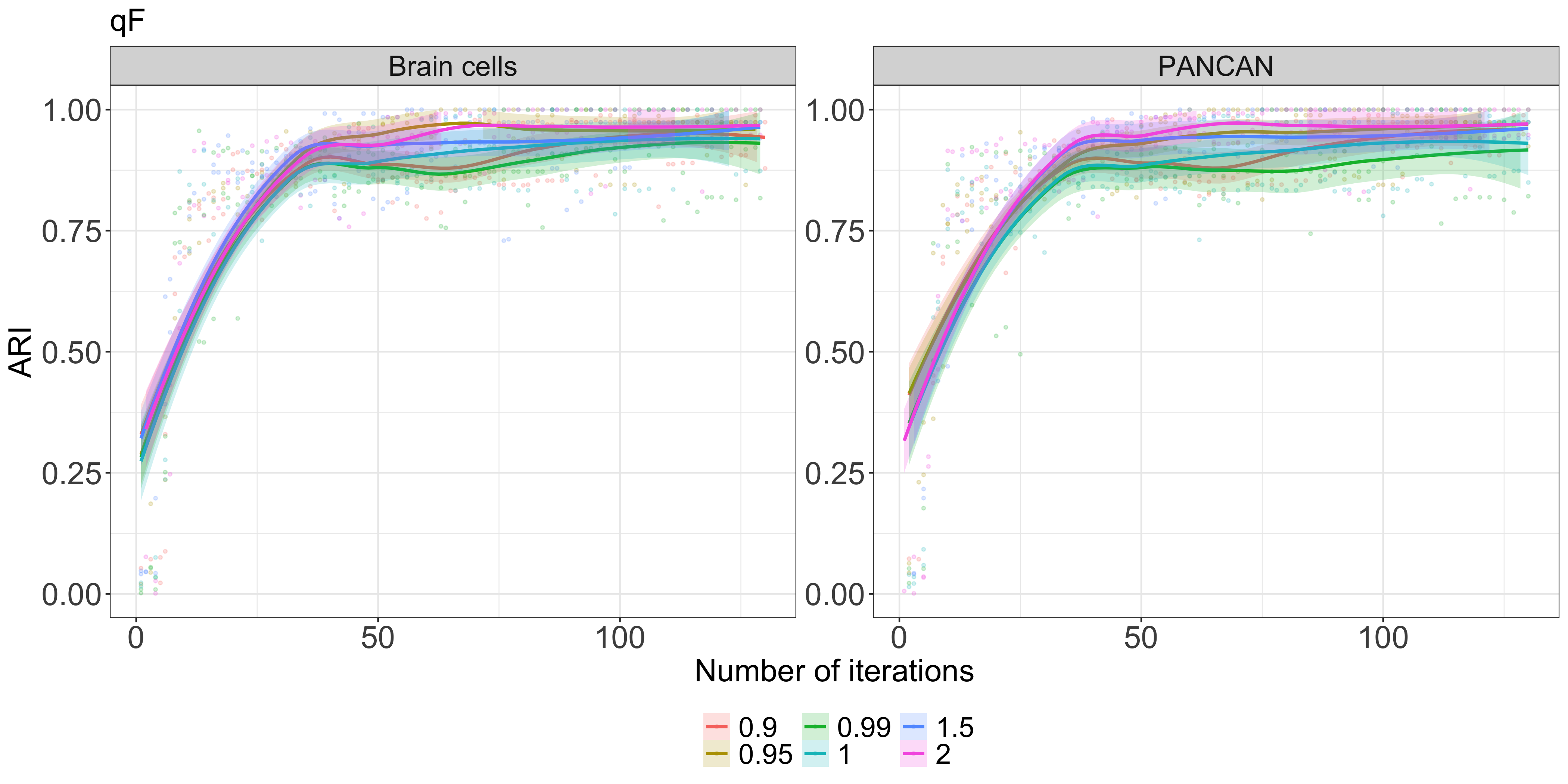}
\caption{Clustering performance measured by ARI with different values of high importance cutoff $\{\tau\}$ in IMPACC using brain cells and PANCAN data. Clustering accuracy is not significantly different with various $\{\tau\}$ settings.}\label{fig:hyper6}
\end{figure}

\begin{figure}[!h]
\centering\includegraphics[scale=0.08]{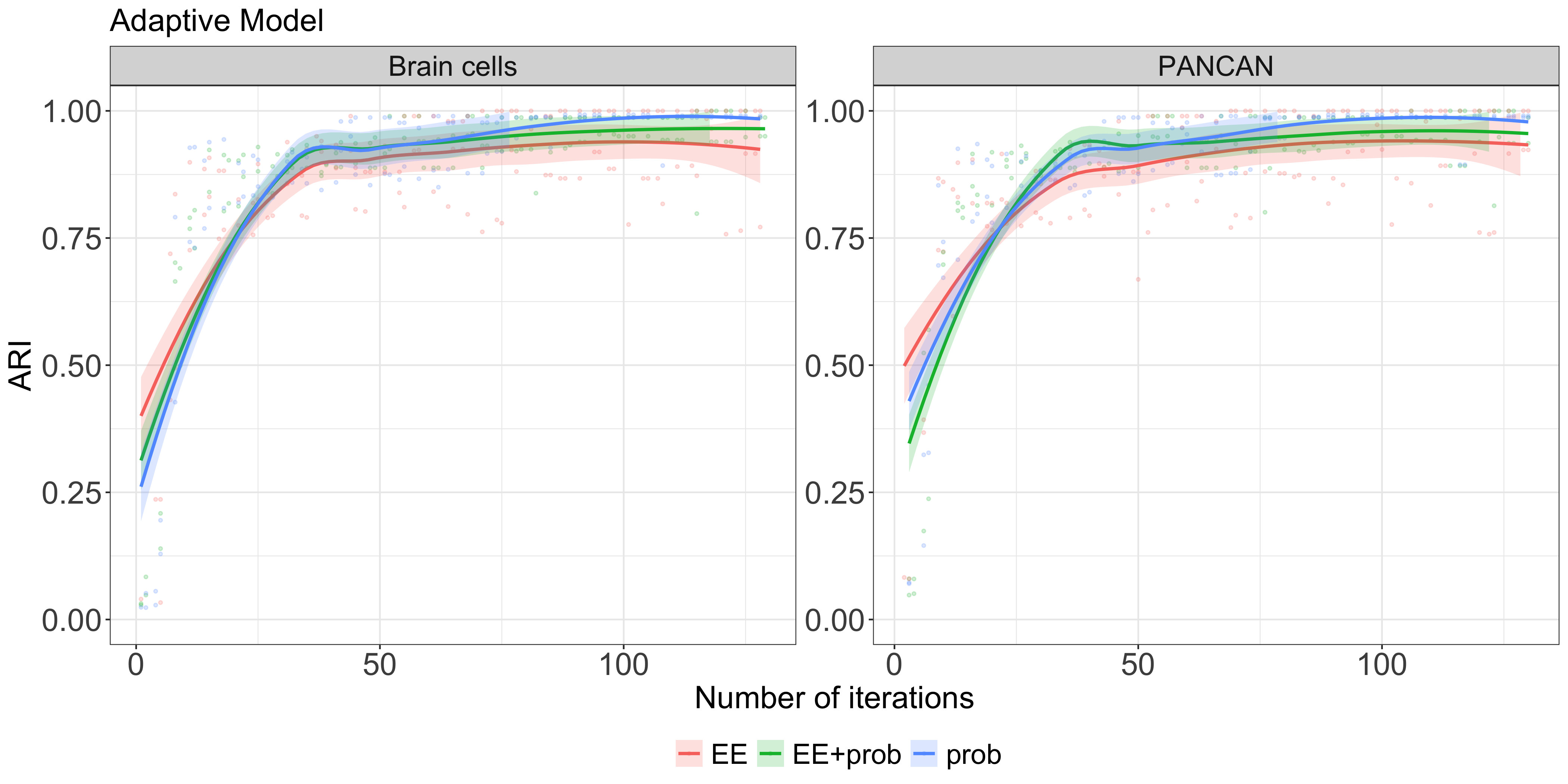}
\caption{Clustering performance measured by ARI with different adaptive sampling schemes in IMPACC using brain cells and PANCAN data. The schemes includede are probabilistic sampling scheme ($prob$), proposed $EE+Prob$ scheme, and $EE$ scheme from [29]. Clustering accuracy is not significantly different with different adaptive sampling methods.}\label{fig:hyper7}
\end{figure}

\begin{figure}[!t]
\centering\includegraphics[scale=0.08]{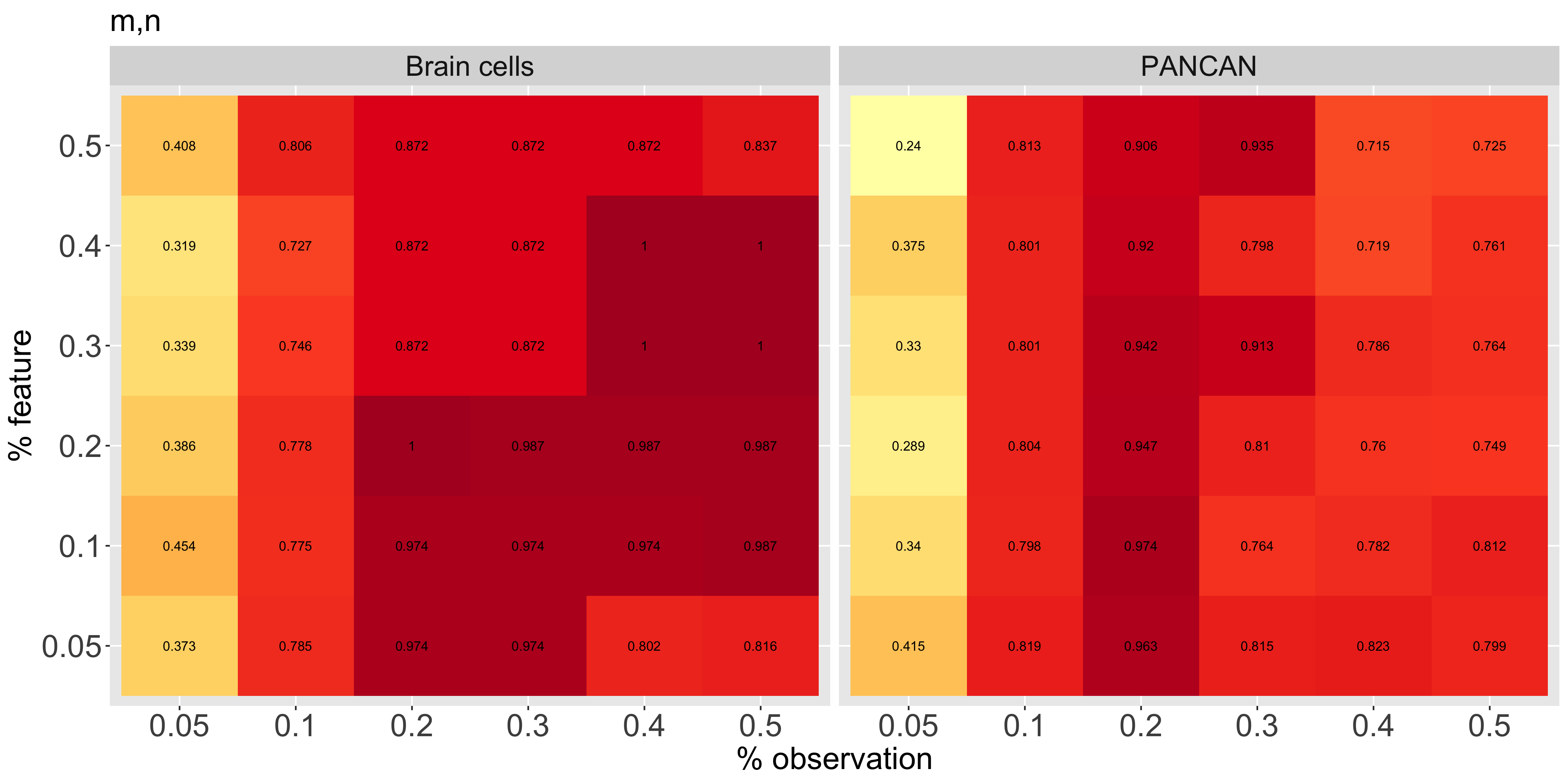}
\caption{Clustering performance measured by ARI with different values of minipatch size $m$ and $n$ in IMPACC using brain cells and PANCAN data.}\label{fig:hyper8}
\end{figure}

There are 3 hyper-parameters in MPCC and 5 additional hyper-parameters in IMPACC. However, only minipatch size related parameters $n$ and $m$ can significantly influence running time and model accuracy. Therefore, choosing a pair of $n$ and $m$ that is relatively small while maintaining good performance would be critical. We find that our framework is fairly robust by setting $m = 0.1$ and $n = 0.25$ in empirical studies. In addition, the remaining hyper-parameters do not need to be tuned, because the model performance is either insensitive and stable to a wide range of parameter choices, or is optimal with one universal value in all cases. Since our models are quite robust and can generate satisfying performance with default parameter settings, we can remarkably reduce computational costs. Hyper-parameters and their default values are summarized in Table~\ref{table:rand_para},~\ref{table:wf_para},~\ref{table:wi_para}, and comparisons on learning accuracy versus number of iterations for different levels of hyper-parameters on two real genomics data are specified in Figure~\ref{fig:hyper1}-\ref{fig:hyper8}.

Besides, We propose a data driven way to conduct parameter tuning based on the consensus matrix. For example, in the selection of minipatch size, we found that our models can yield optimal learning performance by choosing the minimum $m$ and $n$ pair such that the maximum confusion value of the final consensus matrix is less than $0.01$.

\begin{table}[h!]
\centering
 \begin{tabular}{p{2cm} p{6cm} p{2cm} p{1cm}}
 \bf Parameter & \bf Description & \bf Default value & \bf Range \\
\toprule
\bf $m$ & Minipatch features size & $0.1$ &$(0,1]$ \\
\hline
\bf $n$ & Minipatch observation size & $0.25$&$(0,1]$\\
\hline
\bf $h$ & Cutoff quantile on hierarchical tree height & $95\%$  &$(0,1]$\\
\bottomrule
\end{tabular}
\caption{Hyper-parameters in MPCC framework (Algorithm~1).}\label{table:rand_para}
\end{table}

\begin{table}[!h]
\centering
 \begin{tabular}{p{2cm} p{6cm} p{2cm} p{1cm}}
 \bf Parameter & \bf Description & \bf Default value & \bf Range \\
\toprule
\bf $\{\eta\}$ & p-value percentile cutoff & 5\%&$[0,1]$  \\
\hline
\bf $\alpha_F$ & learning rate (feature) & 0.5&$[0,1]$\\
\hline
\bf $\{\tau\}$ & high importance cutoff ($mean$+$\tau sd$) & 1 & $[0,\infty)$\\
\bottomrule
\end{tabular}
\caption{Hyper-parameters in adaptive feature sampling scheme.}\label{table:wf_para}
\end{table}

\begin{table}[!h]
\centering
 \begin{tabular}{p{2cm} p{4cm} p{4cm} p{1cm}}
 \bf Parameter & \bf Description & \bf Default value& \bf Range  \\
\toprule
\bf $\alpha_I$ & learning rate (observation)& 0.5&$[0,1]$\\
\hline
\bf $\{\theta\}$ & high uncertainty cutoff & 95\% weight quantile  & $[0,1]$\\
\hline
\end{tabular}
\caption{Hyper-parameters in adaptive observation sampling scheme.}\label{table:wi_para}
\end{table}

